\documentclass[final,5p,times,twocolumn]{elsarticle}
\usepackage{booktabs}
\usepackage{moreverb,algorithmic}
\usepackage{amsmath}
\usepackage{setspace}
\usepackage{amssymb,lscape}
\usepackage{tabularx}
\usepackage{subcaption}
\usepackage[ruled]{algorithm2e}
\usepackage{graphicx}%
\usepackage{times}
\usepackage{epstopdf}
\usepackage{balance}
\usepackage[x11names]{xcolor}
\usepackage{array,multirow}
\usepackage[T1]{fontenc}
\usepackage{makecell}
\usepackage{eucal}
\usepackage{blindtext}
\usepackage{floatrow}
\newcommand{\newcolorlabel}[2]{%
  \expandafter\newcommand\csname #1\endcsname[1]{%
    \colorbox{#2}{\color{white}\textsf{\textbf{##1}}}}%
}

\newcommand{\newcommenter}[2]{%
  \expandafter\newcommand\csname #1\endcsname[1]{%
    \fcolorbox{#2}{#2}{\color{white}\textsf{\textbf{#1}}}
    {\color{#2}##1}}%
  \expandafter\newcommand\csname at#1\endcsname{%
    \fcolorbox{#2}{#2}{\color{white}\textsf{\textbf{@#1}}}
    {\color{#2}}}%
  \expandafter\newcommand\csname #1hl\endcsname[2]{%
    \colorbox{#2}{\color{white}\textsf{\textbf{#1}}}\fcolorbox{#2}{white}{##2}
    {\color{#2}##1}}%
}

\newcommenter{TODO}{DodgerBlue1}
\newcommenter{rtron}{Green3}

\usepackage{comment}

\usepackage{pdfcomment}

\usepackage{soul}

\usepackage[normalem]{ulem}

\usepackage{csquotes}

\usepackage{units}

\usepackage[font={scriptsize},tableposition=top]{caption}
\usepackage{mathtools}

\DeclarePairedDelimiter{\nint}\lfloor\rceil
\newfloatcommand{capbtabbox}{table}[][\FBwidth]

\newcommand*{\myfont}{\fontfamily{lmtt}\selectfont}

\usepackage{wrapfig}
\usepackage{xkeyval}%

\newwrite\authorbibfile%

\AtBeginDocument{%
  \immediate\openout\authorbibfile=\jobname.aub%
}%

\AtEndDocument{%
\immediate\closeout\authorbibfile
\InputIfFileExists{\jobname.aub}{}{}
}%

\makeatletter

\define@key{authorbib}{scale}[1]{%
\def\AuthorbibKVMacroScale{#1}%
}

\define@key{authorbib}{wraplines}[10]{%
\def\AuthorbibKVMacroWraplines{#1}%
}

\define@key{authorbib}{imagewidth}[4cm]{%
\def\AuthorbibKVMacroImagewidth{#1}%
}

\define@key{authorbib}{overhang}[10pt]{%
\def\AuthorbibKVMacroOverhang{#1}%
}

\define@key{authorbib}{imagepos}[l]{%
\def\AuthorbibKVMacroImagepos{#1}%
}

\makeatother

\presetkeys{authorbib}{imagepos=l,imagewidth=4cm,wraplines=15,overhang=20pt}{}

\newlength{\AuthorbibTopSkip}
\newlength{\AuthorbibBottomSkip}
\NewDocumentCommand{\authorbibliography}{+o+m+m+m}{%
  \IfNoValueTF{#1}{%
  }{%
    \setkeys{authorbib}{#1}%
    \immediate\write\authorbibfile{%
      \string\begin{wrapfigure}[\AuthorbibKVMacroWraplines]{\AuthorbibKVMacroImagepos}[\AuthorbibKVMacroOverhang]{\AuthorbibKVMacroImagewidth}^^J
        \string\includegraphics[scale=\AuthorbibKVMacroScale]{#2}^^J
        \string\end{wrapfigure}^^J
    }%
  }%
  \IfNoValueTF{#3}{%
    \typeout{Warning: No author name}%
  }{%
    \immediate\write\authorbibfile{%
      \unexpanded{\vspace{\AuthorbibTopSkip}}^^J
      \string\noindent\relax
      \unexpanded{\textbf{#3}\par}^^J
      \string\noindent\relax
      \unexpanded{#4}^^J%
      \unexpanded{\vspace{\AuthorbibBottomSkip}}^^J
      }%
  }%
}%

\begin{document}
\begin{frontmatter}
\title{Real-Time Navigation for Autonomous Aerial Vehicles Using Video} %

\author[rutgers]{Khizar Anjum}
\ead{khizar.anjum@rutgers.edu}

\author[rutgers]{Parul Pandey}
\ead{parul\_pandey@rutgers.edu}

\author[rutgers]{Vidyasagar Sadhu}
\ead{vidyasagar.sadhu@rutgers.edu}

\author[boston]{Roberto Tron}
\ead{tron@bu.edu}

\author[rutgers]{Dario Pompili}
\ead{pompili@rutgers.edu}

\affiliation[rutgers]{%
  organization={Rutgers University},
  city={New Brunswick, NJ},
  country={USA},}

\affiliation[boston]{%
  organization={Boston University},
  city={Boston, MA},
  country={USA},}

\begin{abstract}
    Most applications in autonomous navigation using mounted cameras rely on the construction and processing of geometric 3D point clouds, which is an expensive process. However, there is another simpler way to make a space navigable quickly: to use semantic information (e.g., traffic signs) to guide the agent. However, detecting and acting on semantic information involves Computer Vision~(CV) algorithms such as object detection, which themselves are demanding for agents such as aerial drones with limited onboard resources. To solve this problem, we introduce a novel Markov Decision Process~(MDP) framework to reduce the workload of these CV approaches. We apply our proposed framework to both feature-based and neural-network-based object-detection tasks, using open-loop and closed-loop simulations as well as hardware-in-the-loop emulations. These holistic tests show significant benefits in energy consumption and speed with only a limited loss in accuracy compared to models based on static features and neural networks. %
\end{abstract}

\end{frontmatter}

\section{Introduction}
Traditional approaches for autonomous navigation are based on accurate geometric knowledge of the environment~\cite{Lav06, principlesOfRobotMotion}
However, these geometric solutions (occupancy grids, point clouds, etc.) could be improved by adding \emph{semantic} information using techniques such as Computer Vision~(CV) techniques. This is especially true for robots working in environments involving close contact with humans (e.g., an indoor environment~\cite{yang2025novel} or in a search and rescue mission~\cite{Sadhu2020mass}, etc.), because such semantic information can be interpretable for both humans and robots, allowing for a richer understanding of the environment by the robots, and could also be used by humans to better understand or anticipate the actions taken by a robot.  As an example, robots, such as drones, can use the same infrastructure that was developed for humans to navigate. This information can be processed using CV techniques. For example, if a robot detects a ``Turn Right" traffic sign, it might mean that the robot should turn right to avoid a corridor in which humans could be present. However, for this approach to navigation to work, it is imperative that the robot can detect semantic information (signs) in \emph{near real time}. This poses a major challenge, as vision algorithms are generally too complex for robots with limited on-board resources. To counteract such limitations, cloud offloading has been proposed in applications such as localization~\cite{Gouveia:TASE15,Mohanarajah:TASE15} and object recognition~\cite{hunziker2013rapyuta,zweigle2009roboearth}. However, such approaches struggle in the face of high latency or limited network connectivity, leading to performance deterioration. This is intolerable, as a delayed result might not be useful to the robot as the conditions may have changed. In the worst case, a delayed response may result in injury (i.e., when drones work in the proximity of humans).

\textbf{Motivation:}
Hence, real-time applications such as autonomous indoor navigation require a better solution than only offloading to a remote server. We propose a lightweight method for executing such algorithms onboard the drone instead of offloading as an alternative. In this way, we get real-time performance and save energy by eliminating complete dependence on the cloud. We do this by running sign-recognition algorithms using local resources if cloud resources are absent or have high latency. Furthermore, since most of the power on drones is already used by rotors to keep them afloat, a small computational load, if used as an alternative to cloud offloading, should not affect mission parameters much. Sign recognition requires detection algorithms from computer vision, which have seen tremendous performance improvement in the last decade; however, this improvement has been made mostly in terms of accuracy and not much in terms of lower-resource usage. 
By ignoring resource usage, there has been a proliferation of implementation strategies that focus only on achieving the highest accuracy possible. To make such solutions more amenable to on-board processing, we propose to use machine learning methods with different levels of approximations that are selected based on the nature of input data (clutter level, lighting level, etc.), enabling both energy and time savings with respect to a standard one-size-fits-all approach. While image enhancement techniques using generative methods~\cite{huang2024low, kucuk2025new} have been proposed to address such input quality issues, these solutions remain impractical for resource-constrained devices.

\textbf{Proposed Approach:}
This work aims to enhance the responsiveness of detection algorithms while maintaining real-time performance. By optimizing execution speed, we can minimize power draw from the drone's battery, thereby maximizing potential flight duration. Our key insight is that computer vision algorithm selection and parameter tuning should be dynamically adapted based on three key factors: the characteristics of incoming visual data, the available computational and power resources on the device, and the minimum accuracy requirements specified by the application.
To achieve this adaptive optimization, we leverage principles from \emph{approximate computing}. Our methodology carefully balances computational approximations to reduce processing time while maintaining detection accuracy within acceptable bounds. The core of our approach is an intelligent decision system that dynamically selects algorithm parameters to achieve "sufficient" accuracy - delivering acceptable detection performance while significantly reducing computational overhead and energy usage. We implement this decision logic using a Markov Decision Process~(MDP) framework to optimize parameter selection across video frame sequences. The MDP approach was chosen for its lightweight computational footprint and its ability to provide transparent, interpretable decision-making behavior.

\textbf{Contributions:} 
Our work advances lightweight machine learning and decision-making in several ways:
\begin{itemize}
    \item We demonstrate the versatility of our methodology across different machine learning paradigms, including traditional feature-based approaches (Histogram of Gradients and Edgeboxes with SVMs) and modern deep learning methods (CNNs).
    \item We validate our framework through comprehensive testing, progressing from static image analysis to dynamic video processing and ultimately to closed-loop control systems.
\end{itemize}
    Our research yielded the following \textbf{key results}:
\begin{itemize}    
    \item Through systematic offline analysis, we identified optimal algorithm parameters that achieve a $42\%$ speedup while preserving $80\%$ detection accuracy.
    \item Testing on standard benchmarks and custom drone-captured footage demonstrated real-time performance at $\unit[30]{fps}$ for $480 \times 270$ resolution, with accuracy degradation under $2\%$ compared to non-optimized parameters.
    \item We verified practical applicability through hardware-in-the-loop simulations, where a resource-constrained drone successfully navigated custom environments using our framework.
\end{itemize}

\textbf{Generalizability:}
Though we focus on object detection as our primary use case, our MDP-based parameter optimization approach represents a general framework applicable to diverse algorithmic domains requiring real-time performance under resource constraints.

\textbf{Article Organization:}
The remainder of this paper is structured as follows: Sect.~\ref{sec:RelatedWork} contextualizes our work within existing literature. Sect.~\ref{sec:DetectionFramework} presents our decision framework, detailing the MDP formulation and its application to object detection. Sect.~\ref{sec:PerformanceEvaluation} describes our experimental methodology and results. Finally, Sect.~\ref{Sect:Conclusion} summarizes our findings and discusses future directions.

\section{Related Work}\label{sec:RelatedWork}
This section examines existing research in mobile and approximate computing for resource-constrained environments, as well as applications of MDPs in computer vision tasks.

\textbf{Mobile Computation Offloading:}
Contemporary mobile applications leveraging object detection, including interactive games, landmark identification systems~\cite{gavalas2014mobile}, and facial analysis tools~\cite{chen2015glimpse}, frequently delegate computationally intensive operations to remote cloud infrastructure. While this approach enables sophisticated functionality on modest hardware, it introduces significant drawbacks for real-time applications - notably increased latency, substantial energy costs for communication, and dependence on centralized infrastructure. The mobile device cloud paradigm~\cite{Shi2012} attempts to address these limitations by distributing computation across nearby devices. However, this strategy faces practical challenges including device availability, network reliability, and the overhead of task decomposition. Our research takes a different approach by introducing approximation techniques that enable efficient local processing.

\textbf{Approximate Computing for Object Detection:}
While computer vision research has made significant advances in object detection accuracy~\cite{10028728, sun2025evolving}, less attention has been paid to optimizing computational efficiency. For instance, Zhang et al.~\cite{zhang2022attention} developed sophisticated attention mechanisms for object detection but did not consider resource utilization. Our work specifically addresses the computational demands of detection algorithms to enable real-time operation on constrained platforms. Previous efforts~\cite{sidiroglou2011managing,baek2010green,sorber2007eon} have explored individual approximation techniques like loop perforation and alternative implementation strategies. We build upon these foundations by developing an integrated framework that optimizes both algorithmic parameters and input processing while accounting for runtime variability.

The use of MDPs in computer vision is well-established~\cite{karayev2014anytime, xiang_learning_2015}, but their application to approximate computing for mobile platforms represents a novel direction. While architectures like MobileNets~\cite{howard2017mobilenets} have made progress in reducing CNN computational requirements, our framework provides broader applicability across both feature-based and neural approaches. Additionally, we validate our system through comprehensive testing that progresses from static images to video processing and ultimately closed-loop control - aspects not addressed by existing lightweight architectures. The real-time detection system presented in~\cite{fu2024gd} focuses primarily on static image analysis, whereas our approach leverages temporal correlations in video data to achieve additional efficiency gains. Similarly, while BlockDrop~\cite{wu_blockdrop_2018} presents innovative techniques for reducing neural network computation, it does not adapt to input characteristics as our system does.

\section{MDP-Based Decision Framework}\label{sec:DetectionFramework}
This section presents our MDP-based framework for optimizing parameters in object detection systems. The framework utilizes an agent that strategically selects actions to maximize expected rewards, where rewards reflect the balance between processing efficiency and detection quality. Section~\ref{subsec:overview} introduces the core MDP architecture and supporting methodologies. Sections~\ref{subsec:featurebaseddetection} and~\ref{subsec:nnbaseddetection} illustrate the framework's implementation for feature-based detection using Histogram of Gradients~(HoG) and neural network approaches using CNNs, respectively.

\begin{figure}[]
\centering
\includegraphics[width=0.98\linewidth]{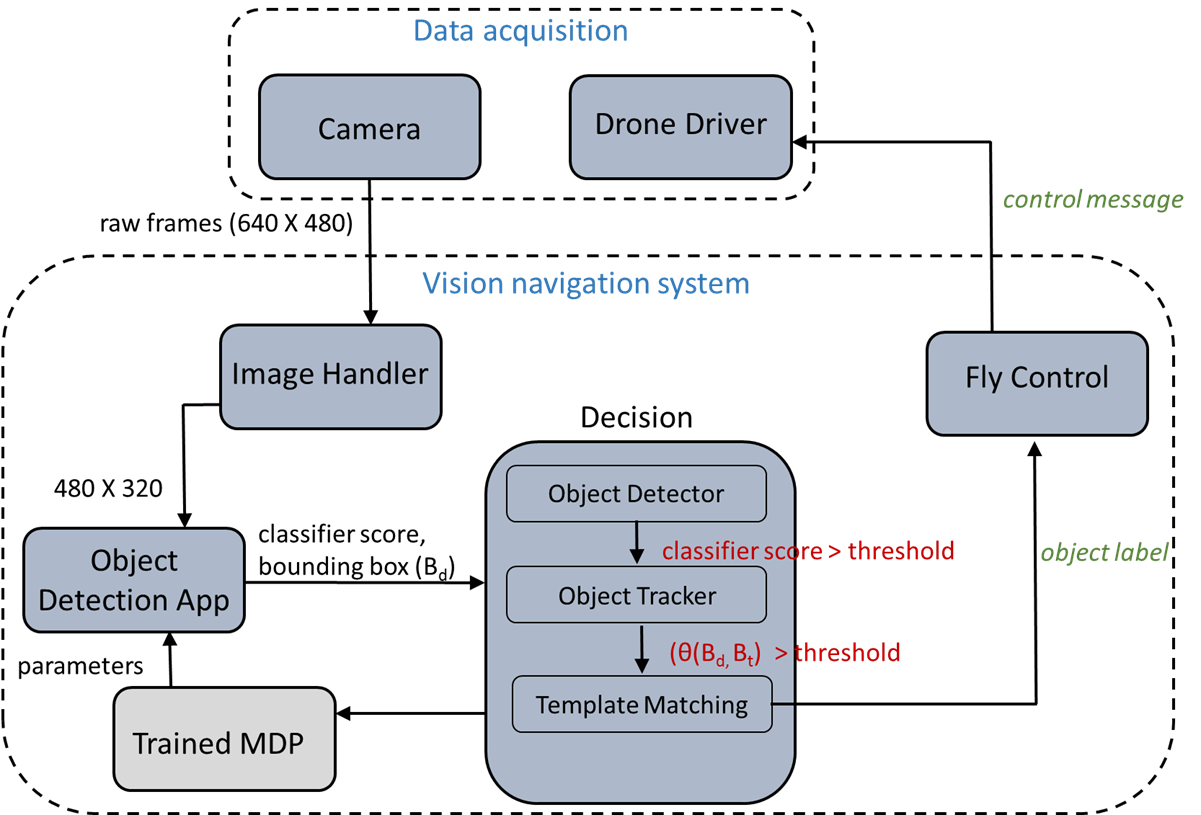}
\caption{System architecture showing key components of our proposed approach. The drone's ROS (master) interfaces with the onboard compute board's ROS (slave, e.g. TX2 or laptop). The decision module processes detected object labels (e.g. 'Stop', 'Turn Left') to generate control commands transmitted to the drone via ROS.}\label{fig:SysD}
\end{figure}

\subsection{Overview}
\label{subsec:overview}
In our framework, incoming video frames are processed by the MDP, which determines suitable object detection algorithm parameters from initial frames and applies them to subsequent frames by utilizing classifier confidence as an accuracy indicator. The MDP re-evaluates parameters when confidence drops below thresholds. Fig.~\ref{fig:SysD} illustrates the complete decision-making pipeline. While the specific MDP implementations differ between feature-based and neural network object detection approaches, we first present the fundamental MDP structure common to both, with implementation details provided in Sections~\ref{subsec:featurebaseddetection} and~\ref{subsec:nnbaseddetection}.

\textbf{Core MDP Architecture:}
The MDP framework comprises four key elements: States, Actions, Rewards, and Policy. The MDP aims to maximize expected reward accumulation through strategic action selection in each state. We detail each component below.

\underline{\emph{States:}}
Each state ${s}\in\mathcal{S}$ represents a configuration tuple containing classifier parameters.

\underline{\emph{Actions:}}
An action $a$ represents the MDP's response to the current state $s$ within its operating environment.

\begin{figure*}[t]
\centering
\begin{subfigure}{0.31\textwidth}
    \includegraphics[width=\linewidth]{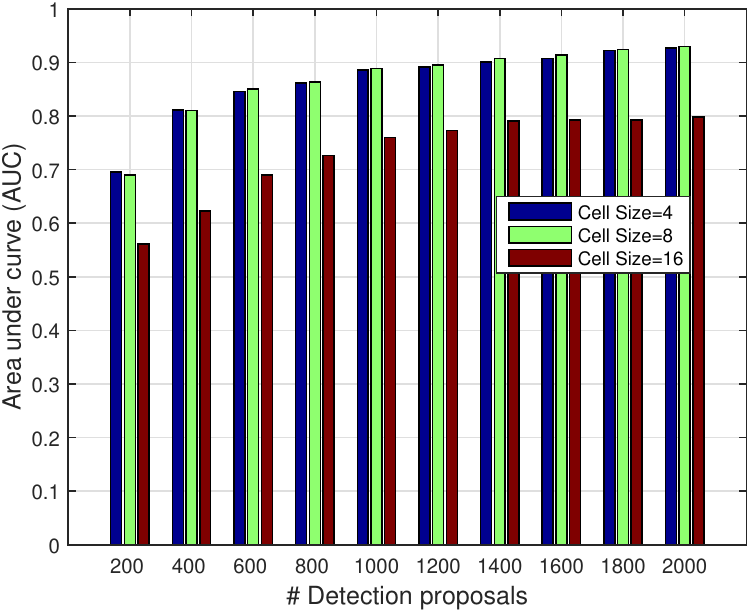}
    \caption{}\label{fig:ApproxHogA}
\end{subfigure}
\begin{subfigure}{0.31\textwidth}
    \includegraphics[width=\linewidth]{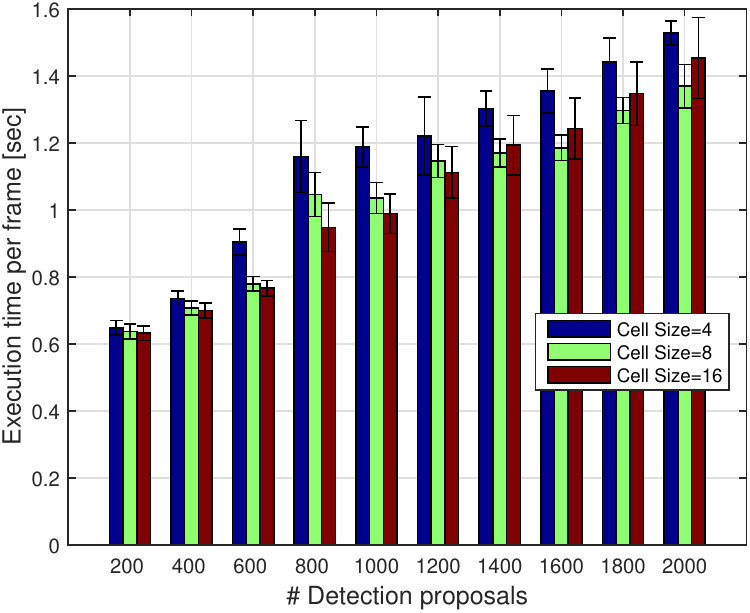}
    \caption{}\label{fig:ApproxHogB}
\end{subfigure}
\begin{subfigure}{0.31\textwidth}
    \includegraphics[width=\linewidth]{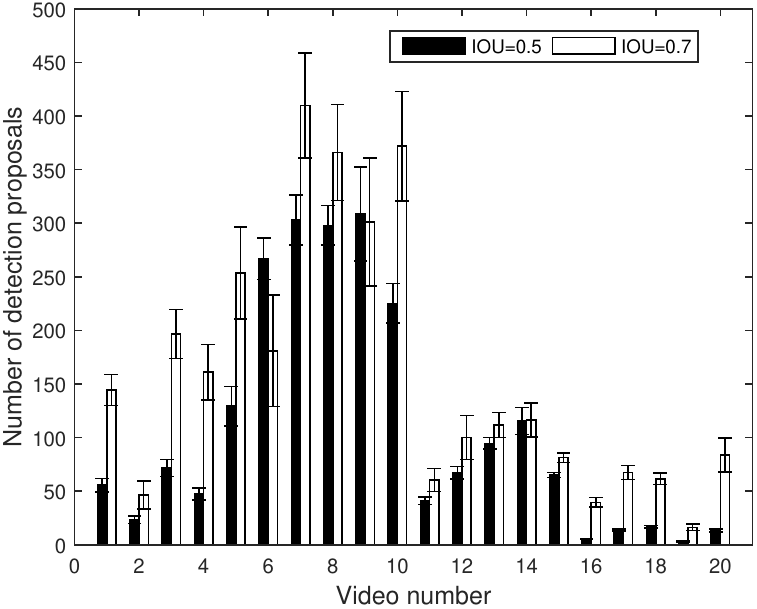}
    \caption{}\label{fig:ApproxHogC}
\end{subfigure}
\vspace{-0.1in}
\caption{Analysis of parameter impacts: (a) Detection accuracy (AUC) and (b) processing time variations across different detection proposal counts and cell sizes in the pipeline from Fig.~\ref{fig:AppAlgo}. (c) Demonstrates that optimal parameters vary significantly across input data, challenging fixed parameter approaches.}\label{fig:ApproxHog}
\end{figure*}

\begin{figure*}[t]
    \centering
    \resizebox{\textwidth}{!}{
    \includegraphics[width=6.7in]{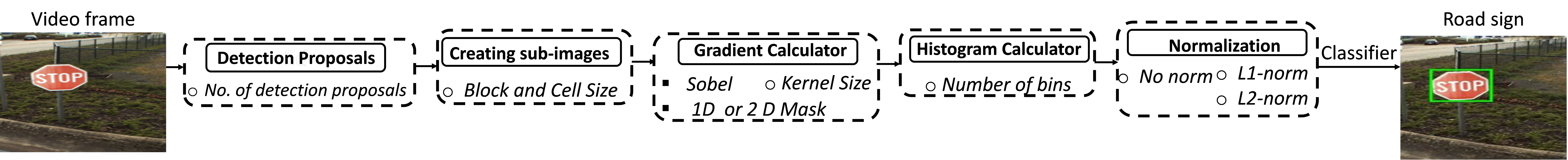} }
    \caption{\scriptsize Feature-based detection pipeline integrating Histogram of Gradients~(HoG) and Edgeboxes algorithms. Dashed blocks indicate modules with multiple implementation variants offering different complexity-performance tradeoffs.}\label{fig:AppAlgo}
\end{figure*}

\underline{\emph{Reward:}}
The reward function $R(s,a,s')$ quantifies the outcome of taking action $a$ in state $s$ and transitioning to state $s'$. Our reward mechanism guides parameter selection by balancing detection quality against computational cost. Since ground-truth accuracy is unavailable during operation, we leverage classifier confidence scores as quality indicators.

\underline{\emph{Policy:}}
A policy $\pi$ defines the mapping that determines action selection $\pi(s)$ for each state $s$. The MDP seeks an optimal policy $\pi^*$ that maximizes cumulative discounted rewards $\Sigma_{t=0}^{\infty} \gamma^t R(s,\pi(s),s')$, where $\gamma \in [0,1)$ represents the discount factor. When $\gamma=0$, the agent prioritizes immediate rewards, while $\gamma$ approaching 1 indicates greater emphasis on long-term rewards.

\textbf{Object Tracking Integration:}
We incorporate temporal information through the Kanade-Lucas-Tomasi Feature Tracker~(KLT), which generates successive bounding boxes $B_t$ based on detector outputs $B_d$. We evaluate tracking quality using Intersection over Union~(IoU):
\begin{equation}
\theta(B_d, B_t) = \frac{B_d \cap B_t}{B_d \cup B_t}.
\end{equation}
The MDP re-evaluates parameters when either classifier confidence falls below threshold $T_c$ or bounding box overlap $\theta(B_d, B_t)$ drops below $\theta_{th}$.

\subsection{Feature-Based Object Detection}\label{subsec:featurebaseddetection}
Here we detail the application of our framework to feature-based detection systems.

\textbf{Model Architecture:}
Our implementation leverages Histogram of Gradients~(HoG) and Edgeboxes algorithms, though the framework accommodates other parameterized detection approaches.

\underline{\emph{Edgeboxes:}}
This component identifies potential object regions (detection proposals) through rectangular bounding boxes, operating independently of specific object categories. Fig.~\ref{fig:ApproxHog} demonstrates how proposal requirements vary with scene complexity, illumination, and viewpoint.

\underline{\emph{Histogram of Gradients~(HoG):}}
HoG extracts discriminative features from proposed regions for object classification. The pipeline comprises multiple stages shown in Blocks 2-4 of Fig.~\ref{fig:AppAlgo}.

\underline{\emph{Classifier:}}
We employ Support Vector Machine~(SVM) classification on HoG features, with model parameters $w$ (weights) and $b$ (bias) trained per object category. The classifier generates confidence scores $c = w^T X + b$ for features $X$ extracted from proposals. We establish confidence thresholds through training data analysis.

\begin{figure*}[t]
    \centering 
    \begin{subfigure}{0.31\linewidth}
        \includegraphics[width=\linewidth]{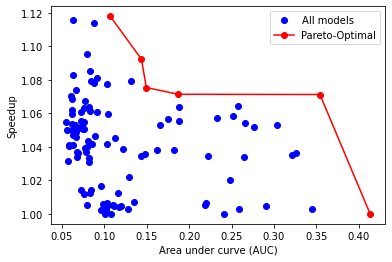} 
        \caption{Performance-complexity tradeoffs for feature-based detection approaches.}\label{fig:model_comp}
    \end{subfigure}
    \begin{subfigure}{0.31\linewidth}
        \includegraphics[width=\linewidth]{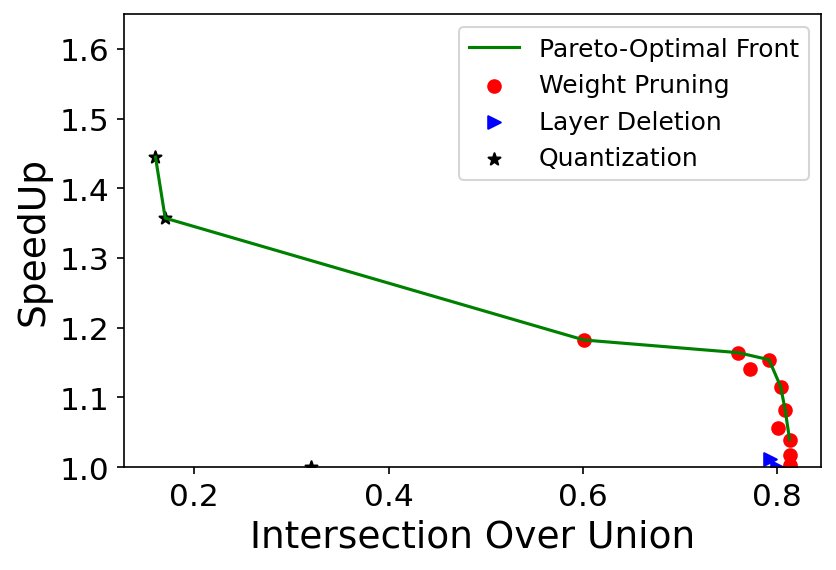} 
        \caption{Single compression technique analysis for neural network detection.}\label{fig:pareto_all}
    \end{subfigure}
    \begin{subfigure}{0.31\linewidth}
        \includegraphics[width=\linewidth]{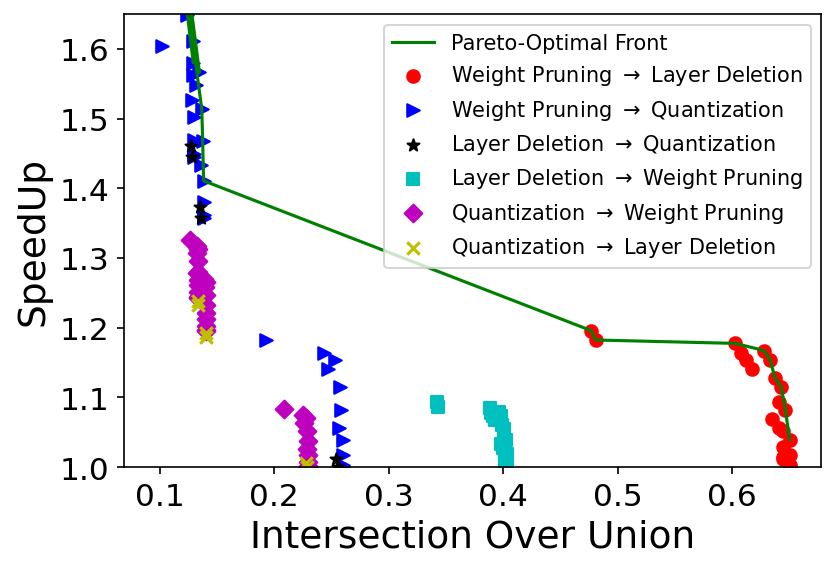} 
        \caption{Combined compression technique analysis for neural networks.}\label{fig:pareto_last3}
    \end{subfigure}
\caption{Comparative analysis of accuracy-speedup tradeoffs across different detection approaches and optimization strategies.}
\label{fig:ApproxActions-1}
\end{figure*}

\textbf{Approximation Methodology:}
Each pipeline stage (Fig.~\ref{fig:AppAlgo}) consists of configurable functions and parameters. Our framework dynamically adjusts these elements based on input characteristics, as evidenced by the varying parameter requirements shown in Fig.~\ref{fig:ApproxHog}.

\textbf{MDP Implementation:}
Below we detail the specific MDP components for feature-based detection.

\underline{\emph{States:}}
States ${s}\in\mathcal{S}$ comprise tuples $\mathcal{S}$ $\doteq$ \{{\myfont c}, {\myfont params}\}, where {\myfont c} represents classifier confidence and {\myfont params} includes detection parameters \{{\myfont BB}, {\myfont bsize}, {\myfont csize}, {\myfont nhist}\} covering proposal count, block size, cell size, and histogram bins. Confidence scores are discretized into three categories using thresholds $c_{opt}$, $c_{ge}$, and $c_{ua}$ to maintain finite state space.

\underline{\emph{Actions:}}
The action space encompasses parameter selection decisions for incoming frame processing.

\underline{\emph{Reward:}}
We define confidence categories (optimal: $c_{opt}$, adequate: $c_{ge}$, inadequate: $c_{ua}$) with associated weights ($w_{opt}$, $w_{ge}$, $w_{ua}$). The reward function is:
\begin{equation}\label{eq:reward}
R(s',w) =\frac{w} {r(s')^{\alpha}}
\end{equation}
where $w$ is assigned based on confidence thresholds, $r(s')$ represents relative computational cost versus minimum-time configuration, and $\alpha$ enables tuning. We empirically set $c_{opt}=0.9$, $c_{ge}=0$, and $c_{ua}=-0.5$, with $w_{ge} \gg w_{opt}$ to encourage efficient parameter selection.

\subsection{Neural Network-Based Object Detection}\label{subsec:nnbaseddetection}
We explore how our MDP decision framework can be applied to neural network-based detection and tracking systems.

\textbf{Model Description:}
When adapting our online selection approach to convolutional neural networks (CNNs), we must address two key differences:
(i)~The state space encompasses network hyperparameters like layer count and kernel dimensions, and (ii)~Novel optimization methods are needed for computational efficiency. Since neural network parameters cannot be modified after training, we must prepare multiple pre-trained model variants beforehand. Below we detail optimization strategies for deploying these networks on resource-limited aerial platforms.

\underline{\emph{Object Detection CNNs:}}
Many pre-trained object detection networks are readily available through frameworks like TensorFlow and PyTorch. The TensorFlow Model Zoo provides networks trained on datasets like COCO~\cite{DBLP:journals/corr/LinMBHPRDZ14}, Kitti~\cite{Geiger2012CVPR}, and Open Images~\cite{OpenImages}. For our evaluation, we utilize Detectron2~\cite{wu2019detectron2} with Faster R-CNN~\cite{DBLP:journals/corr/RenHG015} as the backbone architecture. While our methodology generalizes to other networks, we selected this framework as it represents current state-of-the-art detection capabilities and supports evaluating multiple optimized models. The 101-layer Faster R-CNN backbone particularly demonstrates the benefits of our approach given its substantial computational requirements.

\begin{figure}[t]
\centering
\includegraphics[width=\linewidth]{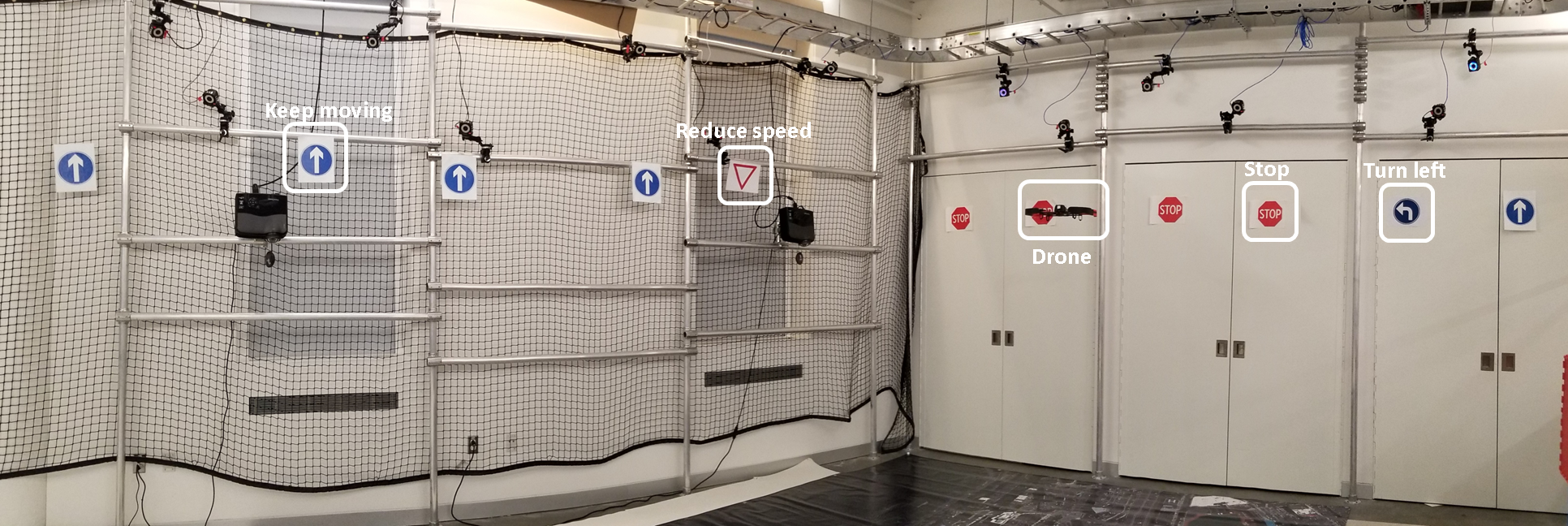}
\caption{Laboratory arena configured as a warehouse environment for collecting the BU-RU dataset. Traffic signs provide navigational guidance for autonomous drone operation.}\label{fig:warehouse_motiv} 
\end{figure}

\begin{figure}[t]
\begin{tabular}{cccccccc}
\includegraphics[width=0.35in]{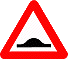}&
\hspace{-0.18in}
\includegraphics[width=0.35in]{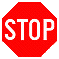}&
\hspace{-0.2in}
\includegraphics[width=0.35in]{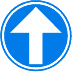}&
\hspace{-0.2in}
\includegraphics[width=0.35in]{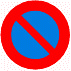}&
\hspace{-0.15in}
\includegraphics[width=0.35in]{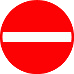}&
\hspace{-0.2in}
\includegraphics[width=0.35in]{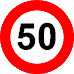}&
\hspace{-0.3in}
\includegraphics[width=0.35in]{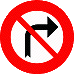}&
\hspace{-0.3in}
\includegraphics[width=0.35in]{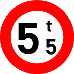}\\
\small A14 & B5 & D1a &E1 & C1 &C43 & C31R & C21
\end{tabular}
\vspace{-0.1in}
\caption{Representative traffic sign samples from the KUL dataset~\cite{mathias2013traffic} with their corresponding dataset identifiers.}\label{fig:Traffic_signs}
\end{figure}

\textbf{Optimization Strategies:}
For neural network deployment, we employ quantization~\cite{Yang_2019_CVPR}, weight pruning~\cite{He_2017_ICCV, 10.1145/3005348}, and layer removal techniques to reduce computational overhead. While these methods help generate efficient network variants, our key contribution lies in using MDP and tracking to dynamically balance accuracy and resource usage.

\underline{\emph{Quantization:}}
This technique converts network weights from floating-point (\texttt{float64}) to integer formats (\texttt{int8}, \texttt{uint8} or \texttt{int32}).
The quantization operation transforms \texttt{float64} tensors $x$ according to:
\begin{equation}
Q(x,scale, zero\_point) = \nint*{\frac{x}{scale} + scale\_point}.
\end{equation}
The $scale\_point$ ensures accurate quantization of zero values to prevent padding-related errors. Each tensor may use different $scale$ values. %
This approach can reduce model size by 75\% while providing 2-4x inference speedup through reduced memory usage and faster integer arithmetic~\cite{krishnamoorthi2018quantizing}.

\underline{\emph{Weight Pruning:}}
Pruning techniques aim to sparsify networks by zeroing weights or removing connections between layers.
We implement global magnitude-based pruning~\cite{blalock2020state} to eliminate a specified percentage of lowest-magnitude weights, reducing memory requirements with minimal performance impact.

\underline{\emph{Layer Removal:}}
This strategy involves selective removal of fully connected or convolutional layers. Removing convolutional layers requires careful consideration of channel counts and dimensional compatibility. We employ an iterative approach - after layer removal, kernel sizes are adjusted to maintain dimensional consistency. Zero-padding or truncation is applied as needed, followed by fine-tuning to preserve performance.

After generating multiple optimized models using these techniques, we identify Pareto-optimal variants by analyzing the accuracy-latency tradeoff. Results from various optimization combinations are presented in Fig.~\ref{fig:ApproxActions-1}.

\textbf{MDP Framework:}
Our MDP training approach mirrors that used for feature-based detection, with adaptations to the state, action and reward definitions as detailed below.

\underline{\emph{States:}}
Each state ${s}\in\mathcal{S}$ comprises the tuple $s \doteq$ \{{\myfont c}, {\myfont Q}, {\myfont $\Omega$}, {\myfont L}\}, where {\myfont c} represents the model's classifier score (categorized using thresholds $c_{opt}=0.9$, $c_{ge}=0$, $c_{ua}=-0.5$), {\myfont Q} indicates quantization type, {\myfont $\Omega$} specifies pruning percentage, and {\myfont L} denotes removed layer count. This state space is pre-computed using Pareto-optimal variants.

\underline{\emph{Actions:}}
The action space involves selecting appropriate detection parameters ({\myfont params}) for processing incoming frames.

\underline{\emph{Reward:}}
We maintain the same reward function~\eqref{eq:reward} used in our feature-based approach.

\begin{figure}
    \centering
    \includegraphics[width=\textwidth]{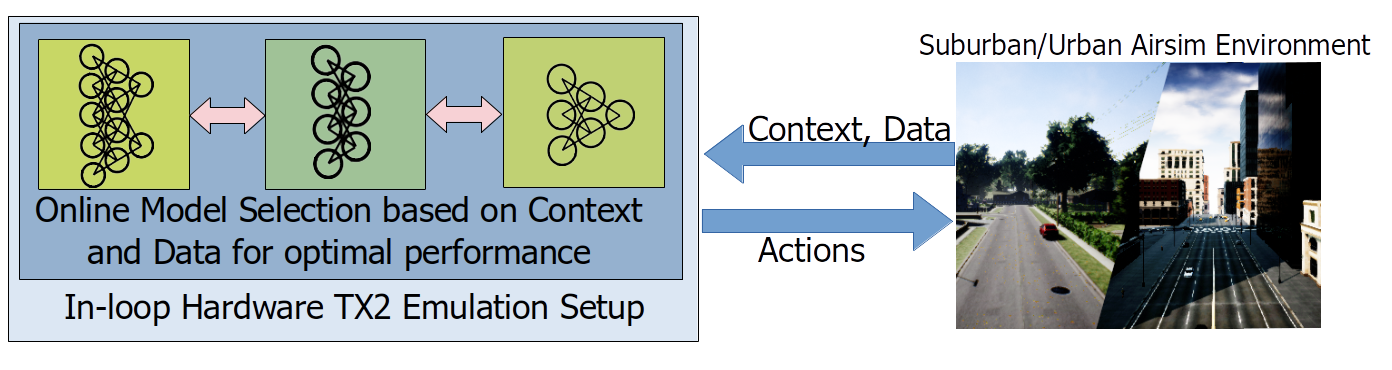}
    \caption{Resource-constrained drone emulation platform utilizing Microsoft's AirSim Simulator~\cite{airsim2017fsr} for hardware-in-the-loop testing.}
    \label{fig:emulator-setup}
\end{figure}

\section{Performance Evaluation}\label{sec:PerformanceEvaluation}
This section evaluates our proposed MDP-based decision framework on publicly available datasets, data collected by our team through a drone platform, and closed-loop hardware emulations. Sect.~\ref{subsec:experimentalsetup} describes the experimental designs and the datasets used to validate our approach. Sect.~\ref{subsec:featurebasedresults} and Sect.~\ref{subsec:nnbasedresults} discuss the results for the feature-based approach and the NN-based approach, respectively.

\subsection{Experimental Setup}\label{subsec:experimentalsetup}
In this subsection, we outline the datasets utilized for evaluation and describe our emulation setup designed to assess the framework's performance under realistic computational constraints.

\begin{figure*}[t]
    \begin{subfigure}{0.32\linewidth}
        \includegraphics[width=\linewidth]{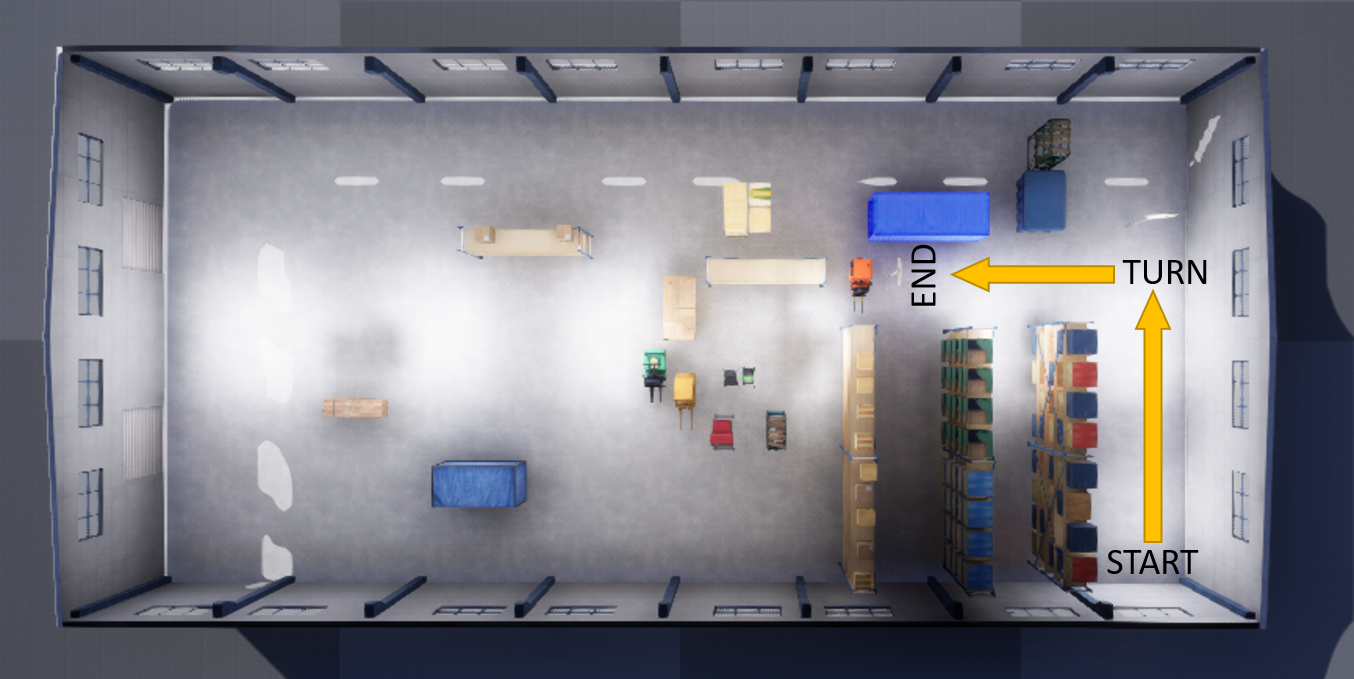} 
        \caption{Easy Maze.}\label{fig:mazesA}
    \end{subfigure}
    \begin{subfigure}{0.32\linewidth}
        \includegraphics[width=\linewidth]{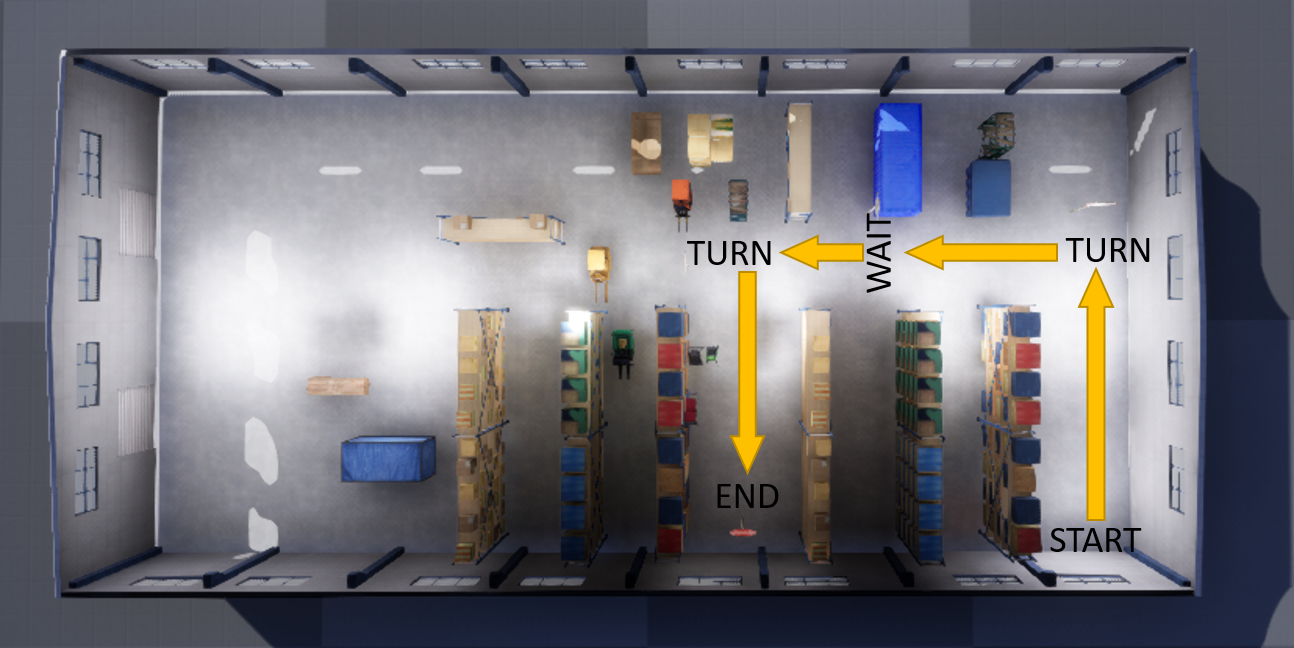} 
        \caption{Medium Maze.}\label{fig:mazesB}
    \end{subfigure}
    \begin{subfigure}{0.32\linewidth}
        \includegraphics[width=\linewidth]{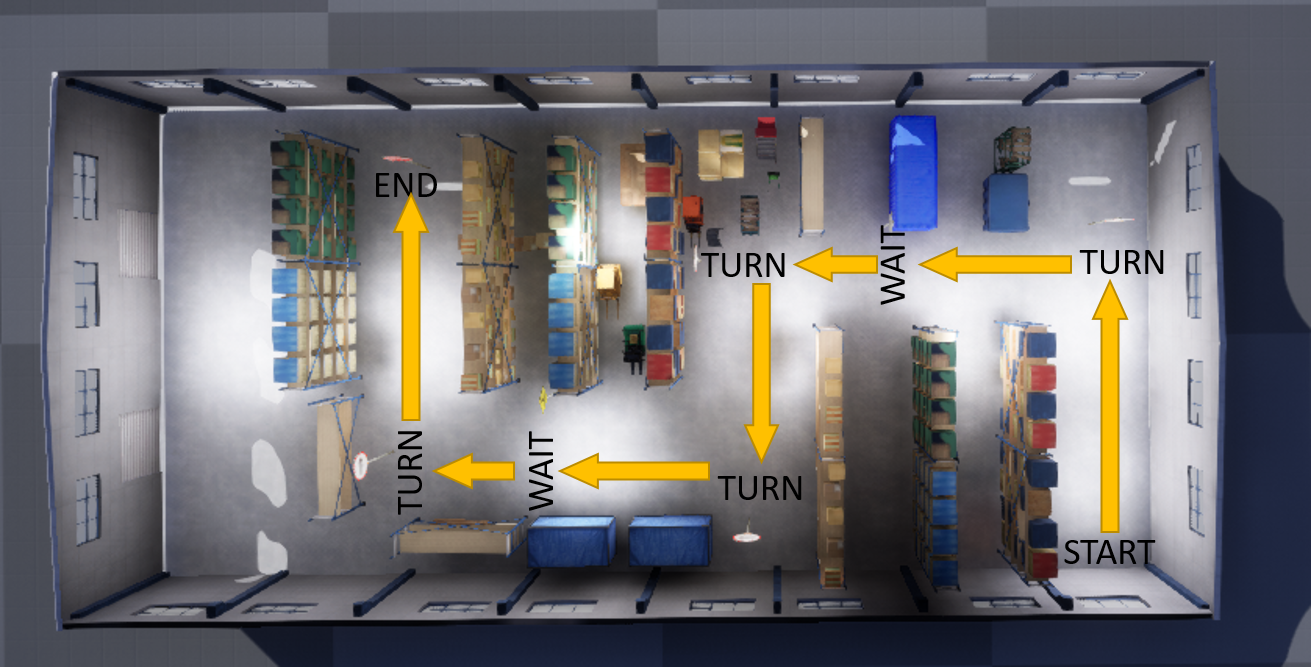} 
        \caption{Hard Maze.}\label{fig:mazesC}
    \end{subfigure}
\caption{Top View: Warehouse mazes created inside AirSim environment. (a)~Easy maze, which is the simplest and has only one turn; (b)~Medium has two turns and has a WAIT (pedestrian) sign; (c)~Hard has four turns and two WAIT signs.}
\label{fig:mazes}
\end{figure*}

\textbf{Datasets:} 
Our evaluation employed two datasets: our custom BU-RU dataset collected using a Bebop drone platform~\cite{Bebop_videos} and the publicly available KUL Traffic Sign Dataset~\cite{mathias2013traffic}.

\emph{\underline{BU-RU Dataset}:} 
This dataset features video footage captured by a drone navigating through warehouse-like environments, as illustrated in Fig.~\ref{fig:warehouse_motiv}. The recordings were made using a Parrot Bebop drone equipped with a 14-megapixel fisheye camera, with traffic signs visible in each frame. Sample traffic signs used are shown in Fig.~\ref{fig:Traffic_signs}. It's worth noting that while we used traffic signs for testing purposes, our approach is adaptable to any specialized signage given appropriate model training. The BU-RU dataset enhances our evaluation by providing a testing environment more closely aligned with our target application. To facilitate comprehensive assessment, we recorded 10 videos each under varying conditions of background clutter (low, medium, high) and illumination levels.

\emph{\underline{KUL Dataset}:}
This dataset consists of videos recorded from a vehicle-mounted camera during road travel. The driving perspective captures traffic signs from various angles and viewpoints, with examples shown in Fig.~\ref{fig:Traffic_signs}. Typically, each video segment spans approximately 40 frames with a single road sign per frame. As the dataset lacked annotations for sign locations and types, we manually annotated a subset for training our algorithms.

\textbf{Emulation Setup:}
To evaluate our approach under realistic conditions, we developed a hardware-in-the-loop emulation platform (outlined in Fig.~\ref{fig:emulator-setup}) using Microsoft's AirSim~\cite{airsim2017fsr} drone simulator and an NVIDIA Jetson TX2 as the computational proxy. The simulator streams live video to the TX2, which processes each frame and returns control commands to the simulated drone. This configuration closely approximates real-world deployment with onboard processing capabilities. While AirSim provides various realistic urban and suburban environments for testing, we also designed custom warehouse-like mazes of increasing complexity to better align with our target application scenario. These mazes, shown in Fig.~\ref{fig:mazes}, include pedestrian signs that require the drone to stop and proceed only when no pedestrians are detected nearby (marked as "WAIT" in the figures). The three maze configurations—Easy (Fig.~\ref{fig:mazesA}), Medium (Fig.~\ref{fig:mazesB}), and Hard (Fig.~\ref{fig:mazesC})—present progressively greater navigational challenges.

\subsection{Feature-Based Object Detection}
\label{subsec:featurebasedresults}
We present here our evaluation methodology and the corresponding results for feature-based object detection on the KUL dataset and the BU-RU dataset, respectively. We also have clearly defined the hyperparameters and the various thresholds used in our approach for the reader's clarity in Table~\ref{tabl:thresh}. Furthermore, since the evaluation of the feature-based object detection is done on a 3.4 GHz Intel i7 CPU, we found it prudent to mention the runtimes experienced by various components of our algorithm: Communication between various components took less than 3 ms, ROS (Robot Operating System) preprocessing took 5.4 ms, MDP parameter search took 36 ms, object tracking took 3.7 ms, and finally the feature extraction part took anywhere between 100--800 ms. However, we should also mention that the results obtained from this computational platform are still relevant for drones, as the relative speedup obtained by using our method is enough to motivate its computational savings.

\begin{figure}[t]
\begin{floatrow}
\resizebox{2.5in}{!}{%
\capbtabbox{%
\begin{tabular}{|c|c|} 
 \hline
 \textbf{Thresholds} & \textbf{Value} \\ [10pt]
 \cline{1-2}
 Re-trigger MDP& 100 frames\\
 \cline{1-2} 
Re-initialize KLT & 10 frames\\
 \cline{1-2} 
Score thresh $T_c$ (KUL) & 0 \\
\cline{1-2} 
 Score thresh $T_c$ (BU-RU) & -1 \\
\cline{1-2} 
Access MDP per image & 5 \\
 \cline{1-2} 
KLT thresh $\theta_{th}$ & 0.5 \\
 \cline{1-2} 
Max. object proposals & 2000 \\
 \cline{1-2}  
IoU for evaluation & 0.5 \\
 \cline{1-2}  
 \hline
\end{tabular}
}
{%
  \caption{{Various thresholds considered in the performance evaluation of our solution.}\label{tabl:thresh}}
}
}
\end{floatrow}
\end{figure}

\begin{figure}
\begin{floatrow}
\resizebox{\columnwidth}{!}{%
\capbtabbox{%
  \captionsetup{font=small}
  \caption{{\textbf{KUL dataset:} Analysis of detection accuracy and computational efficiency for traffic signage identification, comparing baseline configuration~\cite{alexe2012measuring} versus optimized parameter selection through Markov decision process.}\label{table:perf_dataset}}
}
{%
  \begin{tabular}{p{1.00cm} c c c c p{1.00cm} p{1.00cm}}
\toprule
    &       \multicolumn{2}{c}{\textbf{Average}} & \multicolumn{2}{c}{\textbf{Good Enough}} \\
\cmidrule(lr){2-3}\cmidrule(lr){4-5}
    {\textbf{Codes}}  & \textbf{P} & \textbf{E~[s]} & \textbf{P} & \textbf{E~[s]} & {\textbf{{Speed-up~[\%]}}} & {\textbf{{Accuracy Loss~[\%]}}}\\[10pt]

    \hline
    \hline
A14 &   0.8   &1.8~$\pm$~0.3   & 0.7 & 1.6~$\pm$~0.3 & 14.0 & 16.4\\  
B5    & 0.6   &  2.2~$\pm$~0.1 & 0.5  &  2.6~$\pm$~0.5& -  &\\
D1a   	&  1.0 	&  2.1~$\pm$~0.1   & 0.8   &  1.1~$\pm$~0.1 & 40.1 & 22.0\\
E1       & 1.0  &  2.1~$\pm$~0.3   & 0.9   &   2.1~$\pm$~0.1 & 45.7 & 4.0\\
C1     & 0.7 &   2.1~$\pm$~0.2 &0.6 & 1.6~$\pm$~0.1& 22.8  & 6.7\\
C43     & 0.8  &2.1~$\pm$~0.1 & 0.6&  1.9~$\pm$~0.1& 6.7 & 26.8\\
C31R     & 0.8   & 2.0~$\pm$~0.2&0.8 & 1.9~$\pm$~0.7 & 7.4 & 5.0\\
C21  &  0.5  &1.9~$\pm$~0.1 & 0.5& 1.6~$\pm$~0.1 & 17.0 & 0\\
\hline   
\bottomrule
\end{tabular}
}}
\end{floatrow}
\end{figure}

\begin{figure}[t]
    \centering
    \begin{subfigure}{0.47\textwidth}
        \includegraphics[width=\linewidth]{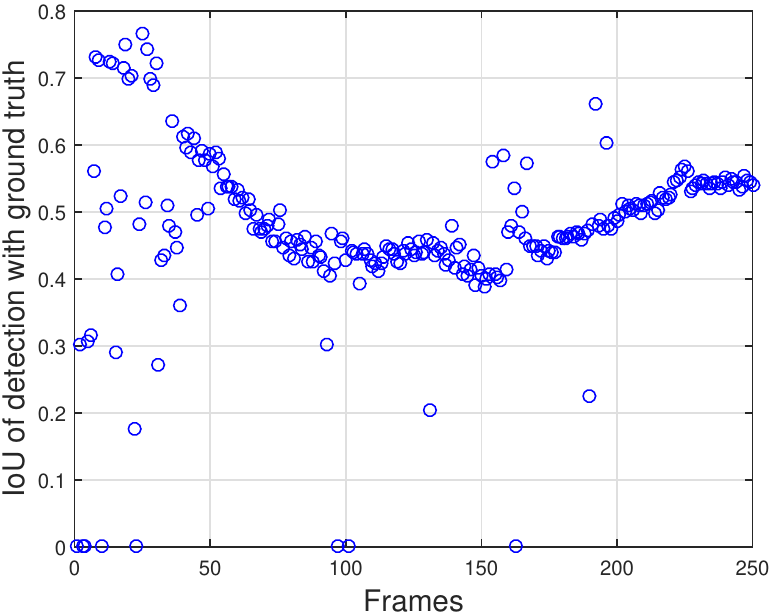} 
        \caption{}\label{fig:IOU_v3}
    \end{subfigure}
    \begin{subfigure}{0.47\textwidth}
        \includegraphics[width=\linewidth, height=1.3in]{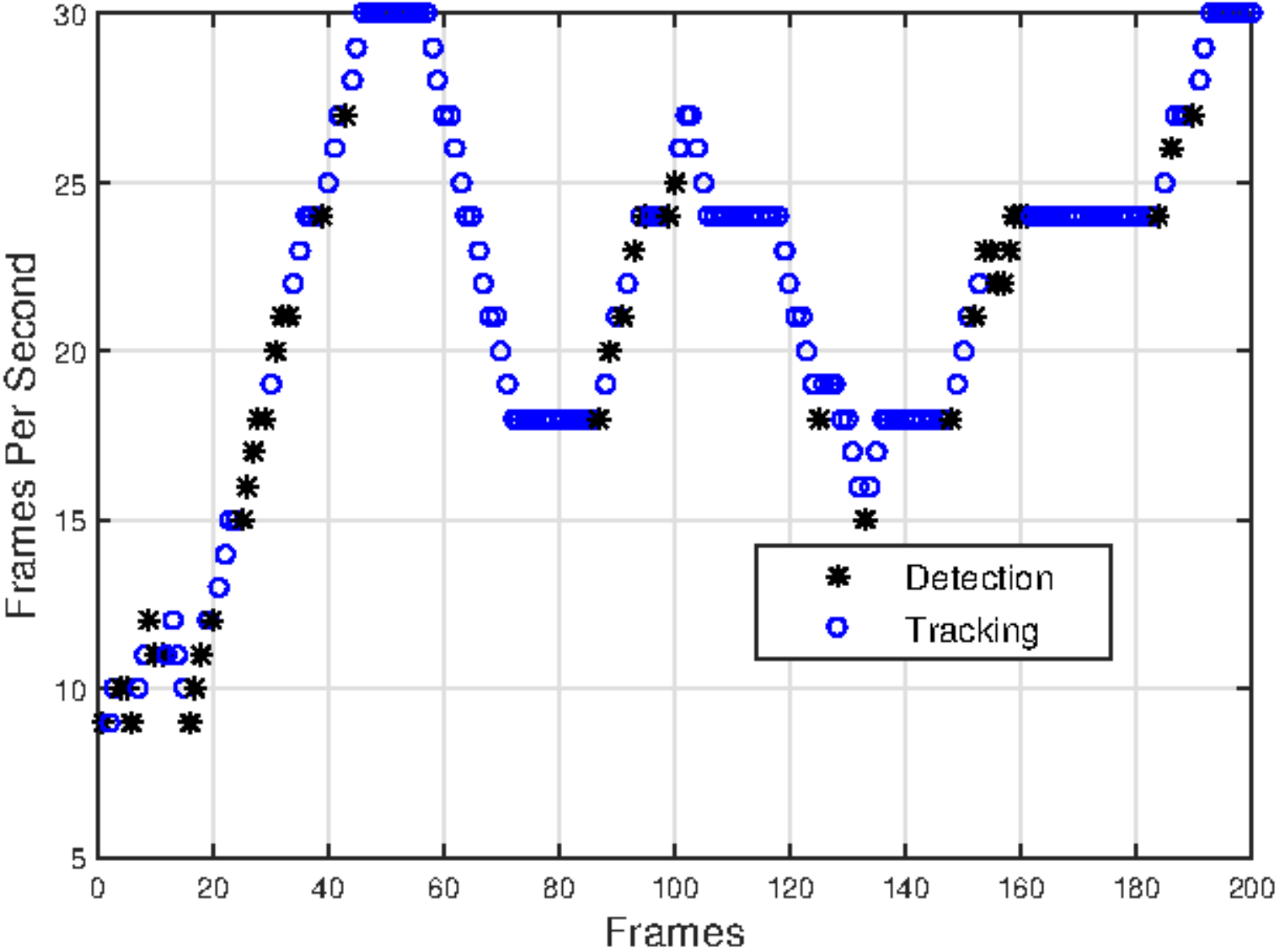} 
        \caption{}\label{fig:MDPTracking}
    \end{subfigure}
    \begin{subfigure}{\textwidth}
        \includegraphics[width=\linewidth, height=1.3in]{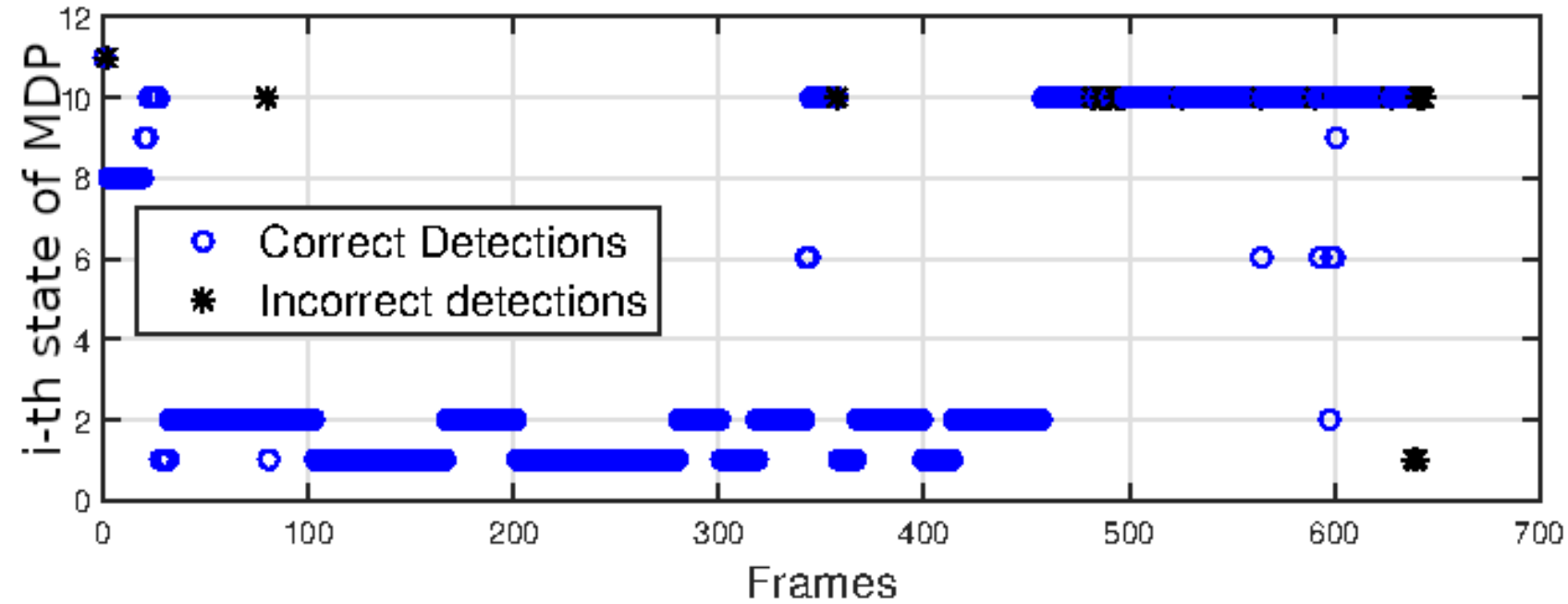} 
        \caption{}\label{fig:MDPStates}
    \end{subfigure}
    \vspace{-0.1in}
\caption{(a)~and~(b)~show real-time performance in terms of IOU and frames per second for feature-based object detection. (c)~Illustration of the MDP state space for the object detection algorithm that the navigation system accesses for an example video sequence from the BU-RU dataset.}
\label{fig:MDPAdaptive_2}
\end{figure}

\begin{figure}[t]
    \begin{subfigure}{0.47\textwidth}
        \includegraphics[width=\linewidth,height=1.3in]{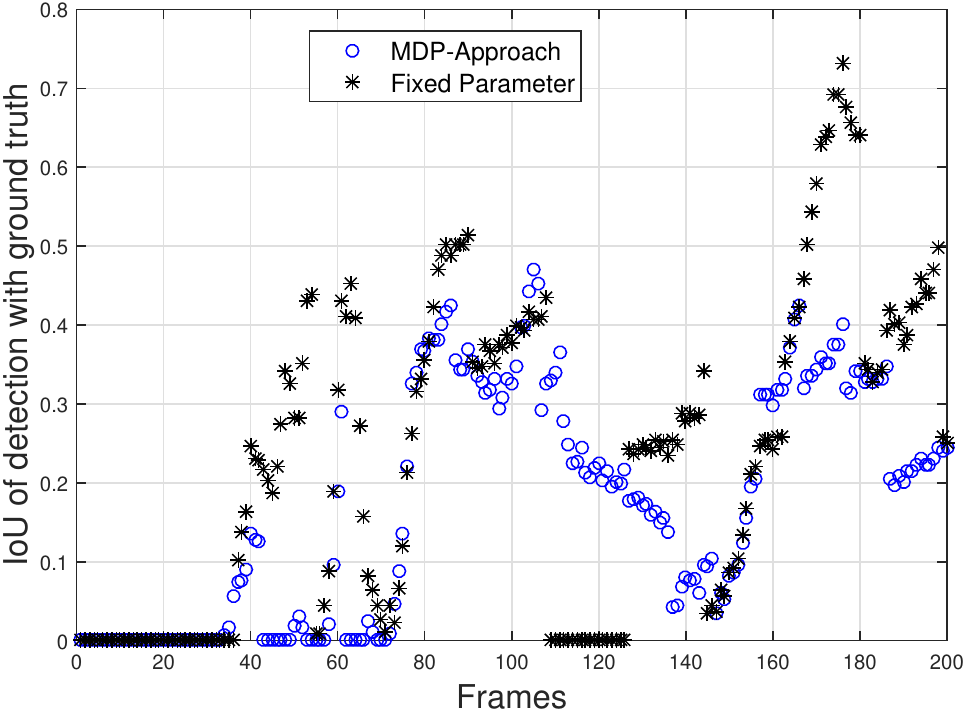} 
        \caption{}\label{fig:MDPAdaptive_1A}
    \end{subfigure}
    \begin{subfigure}{0.47\textwidth}
        \includegraphics[width=\linewidth,height=1.3in]{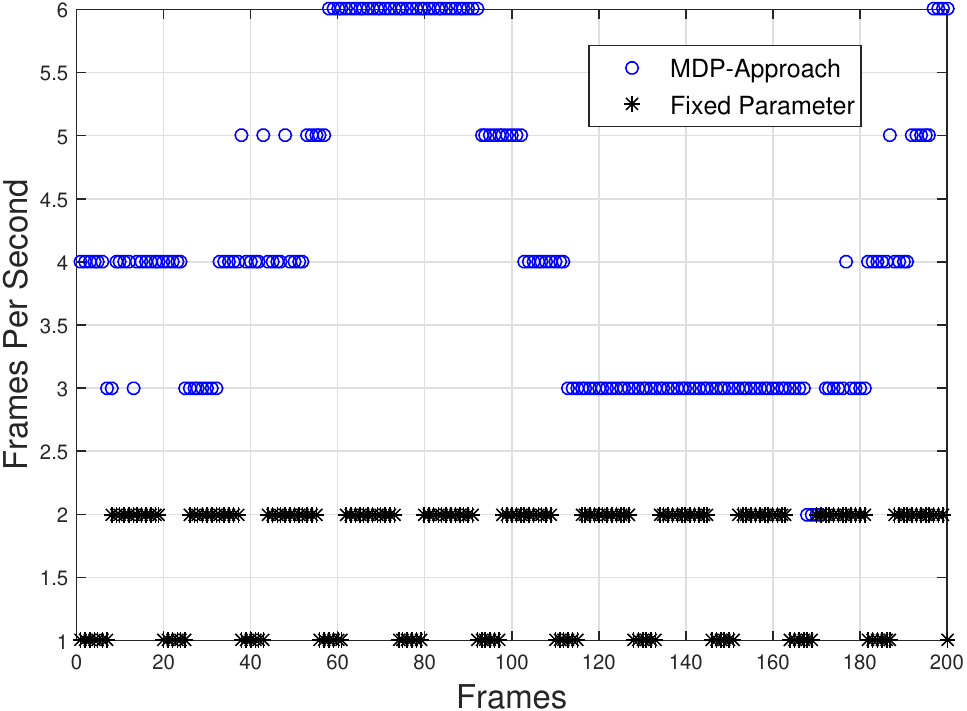} 
        \caption{}\label{fig:MDPAdaptive_1B}
    \end{subfigure}
\vspace{-0.1in}
\caption{(a)~IoU of ground truth with detected locations and (b)~Frames per second achieved for feature-based detection versus fixed-parameter approach on a sample video sequence.}
\label{fig:MDPAdaptive_1}
\end{figure}

\textbf{Performance on the KUL dataset:}
Performance of our MDP-based algorithm for feature-based object detection is in Table~\ref{table:perf_dataset}. The codes refer to the traffic signs shown in Fig.~\ref{fig:Traffic_signs}. The table sums up the improvement in execution time and accuracy loss in terms of percentages. The `Average' column shows the performance and execution time for static parameter selection without MDP-based selection, while the `Good Enough' column shows the performance and execution time for our MDP-based approach. As shown, our algorithm provides great speedups such as $45.7$\%, $40.1$\%, and $22.8$\%.

\textbf{Performance on the BU-RU dataset:}
We evaluate our approach on video sequences from the BU-RU dataset to show the advantage of selecting different parameters adaptively for road sign detection for the HoG object detection algorithm. 
We get a glimpse of the internal workings of our MDP in Fig.~\ref{fig:MDPStates}, where we can see that MDP chooses different states based on the input frames. You will notice that there are contiguous regions where the MDP state does not change for a long time. This is because MDP is not invoked at every frame; instead, the KLT tracker is used to further reduce processing time. The MDP and the detector are invoked if the classifier score is below the threshold $T_m$. This can happen because of the drift of the tracker. 
In Fig.~\ref{fig:MDPTracking}, we show the frames in terms of the operations performed on them, that is, detection or tracking. We can see that detection is performed way less than the tracking operation and plays a role in our speedup gain. This combination allows us to achieve real-time performance as we can see that it approaches 30 fps (frames per second), which is useful for real-time control.

Similarly, in Fig.~\ref{fig:MDPAdaptive_1}, we compare our adaptive approach with a fixed parameter approach. In Fig.~\ref{fig:MDPAdaptive_1B}. It can be seen that a fixed parameter (we use the tuple <1000,2,8,9> as in~\cite{alexe2012measuring}) approach processes a smaller number of frames per second compared to our adaptive MDP-based approach. Please note that this gain is achieved without the incorporation of a KLT tracker and is just meant to be a pure comparison between a fixed-parameter approach and an MDP-based adaptive-parameter approach.

\begin{figure*}[t!]
    \begin{subfigure}{0.31\textwidth}
        \includegraphics[width=\linewidth,height=1.85in]{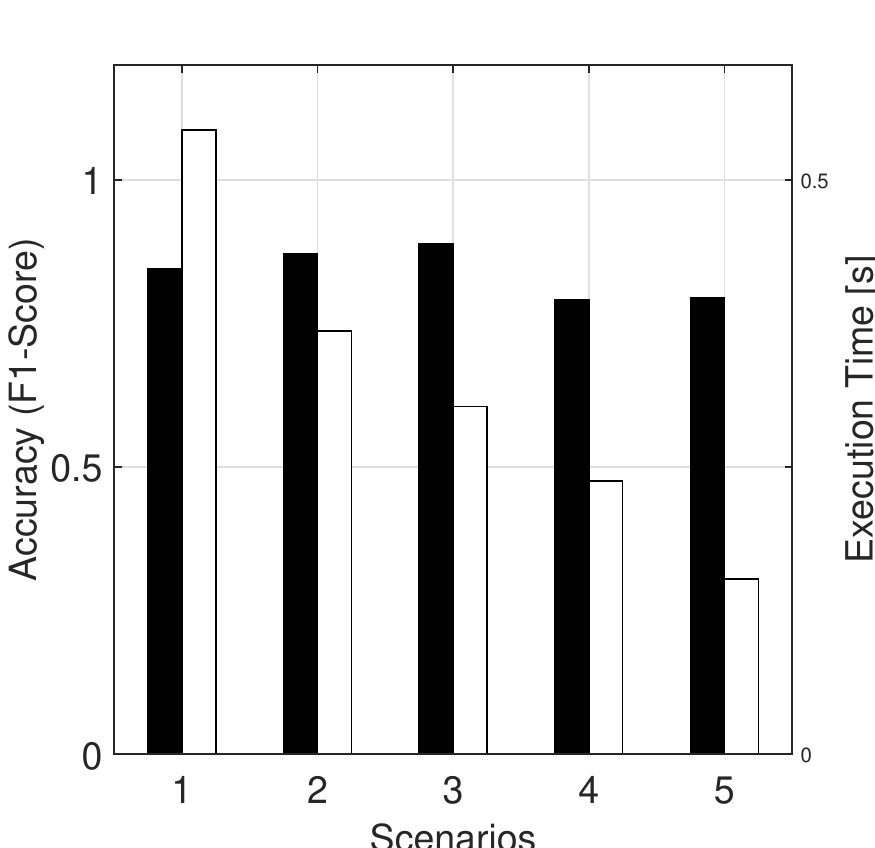}
        \caption{}\label{fig:PerfMDPA}
    \end{subfigure}
    \begin{subfigure}{0.31\textwidth}
        \includegraphics[width=\linewidth,height=1.85in]{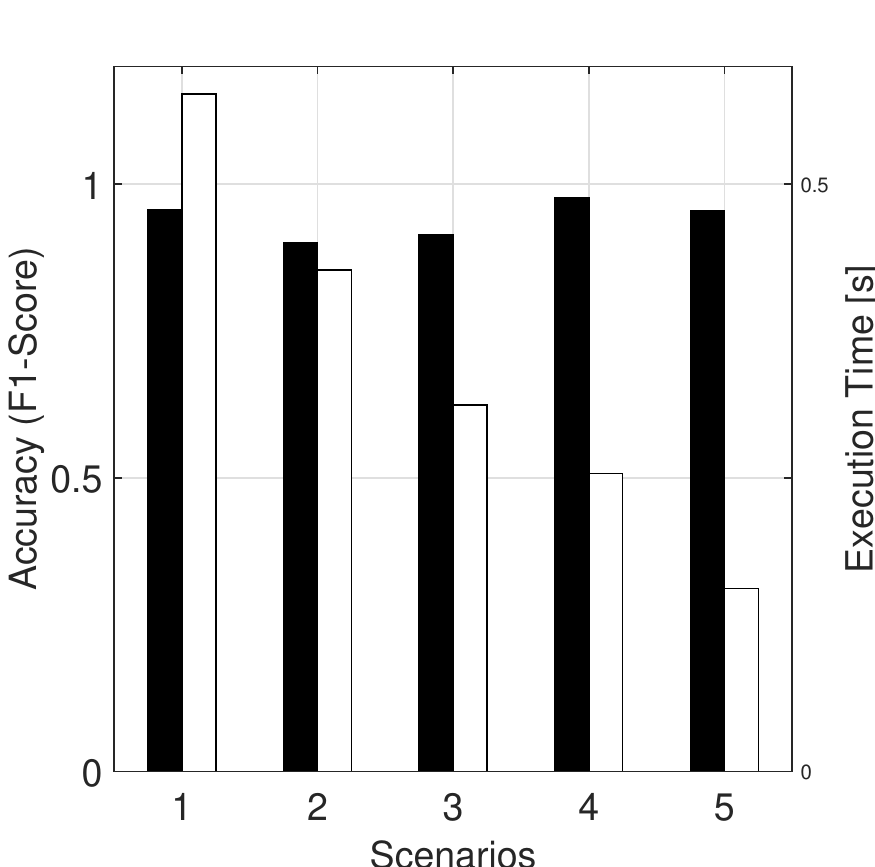}
        \caption{}\label{fig:PerfMDPB}
    \end{subfigure}
    \begin{subfigure}{0.31\textwidth}
        \includegraphics[width=\linewidth,height=1.85in]{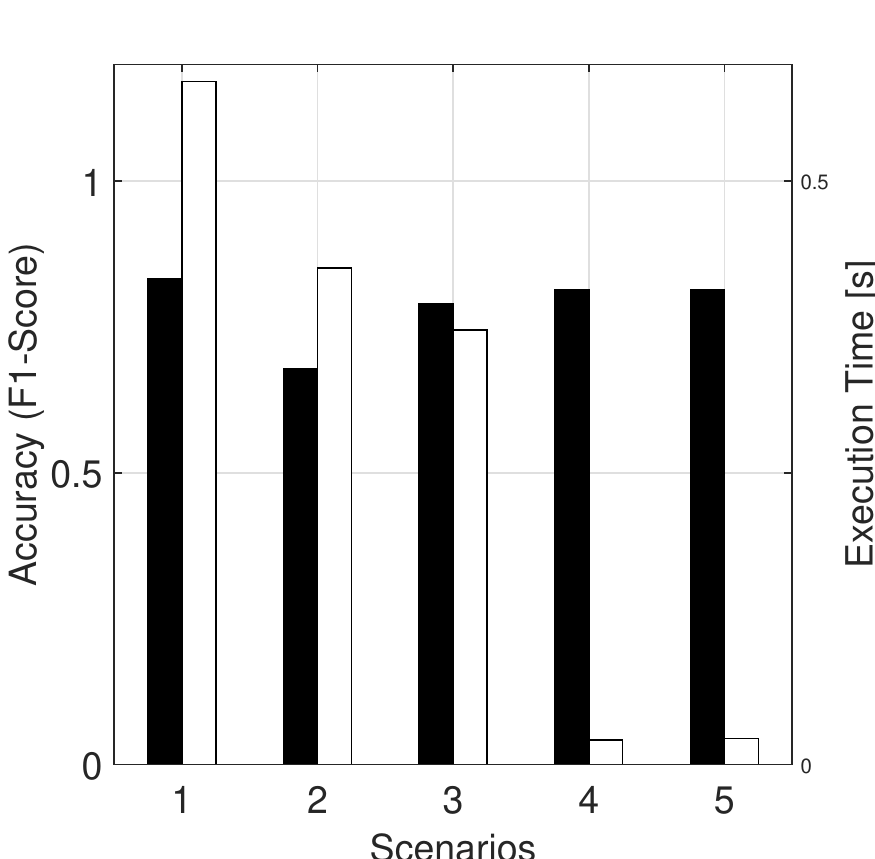}
        \caption{}\label{fig:PerfMDPC}
    \end{subfigure}
\vspace{-0.1in}
\caption{\textbf{BU-RU dataset:} Comparative analysis of detection methodologies across varying conditions: (1)~Static parameter configuration, (2)~Stochastic parameter selection, (3)~MDP-optimized parameter adaptation, (4)~Static parameters with tracker integration, and (5)~Hybrid MDP-tracker framework evaluated under (a)~minimal environmental complexity, (b)~dense scene clutter, and (c)~suboptimal lighting conditions.}
\label{fig:PerfMDP}
\end{figure*}

To evaluate the effectiveness of different parameter selection strategies, we analyze five distinct approaches in Fig.~\ref{fig:PerfMDP}:
(i)~Static parameter detection, where detection uses constant parameters across all frames;
(ii)~Stochastic parameter choice with tracking support, utilizing randomly selected parameter combinations;
(iii)~Adaptive MDP parameter selection, which dynamically adjusts detection parameters through our MDP framework when the intersection-over-union between tracker and detector outputs exceeds threshold $\theta_{th}$;
(iv)~Static parameters with tracking assistance, employing fixed detection parameters but only invoking the detector when tracked region and template overlap falls below threshold $T_m$; and
(v)~MDP-guided adaptive detection with tracking support, which combines tracking with MDP-based parameter optimization.

\emph{\underline{Background Clutter}:}
Fig.~\ref{fig:PerfMDP} shows the results for these different scenarios. In the results for a high background clutter scenario, we see that the fixed-parameter approach (selected from~\cite{alexe2012measuring}) shows the best performance in terms of accuracy (F1 score), but also incurs the most execution time. However, the MDP-based parameter performs $39\%$ better in terms of execution time. In Scenarios~4 and 5, we see that even with the addition of a tracker, our MDP approach further reduces the execution by $35\%$. Similar results are seen for low background-clutter scenarios. This shows that our approach performs well in all these different settings.

\emph{\underline{Poor Illumination}:}
In Fig.~\ref{fig:PerfMDP}(c), we show the MDP performance when illumination is low. Similarly to what we saw in the low and high background clutter scenario, here, again, our MDP-based adaptive parameter approach outperforms the fixed parameter approach and achieves a gain of $35\%$ in terms of execution time. We also noticed that the tracker could perform very well in this scenario, leading to very low execution times when the tracker was introduced in Scenarios~4 and 5. 

\textbf{Comparison with Related Works:}
Xiang et al.~\cite{xiang_learning_2015} use an MDP for multiobject tracking that keeps track of different objects while tracking. The MDP then generates a dynamic similarity score between object detection in image frames in a video that matches the objects together for tracking. Compared to our approach, this is a use of the MDP for purely tracking purposes while we apply a well-known KLT tracker for tracking objects. This KLT is used to achieve further speed-up, with the MDP being triggered only after a set number of frames. Hence, our work applies the MDP to \textit{choose} detector parameters instead of calculating a similarity measure between object detection to achieve speed-up. Furthermore, the work is concerned with increasing the accuracy of object tracking and tackling the disappearance/appearance of targets in a robust manner, but we are concerned with achieving \textit{good enough} performance with the least amount of resources.
Karayev et al.~\cite{karayev2014anytime} uses an MDP to explicitly set a budget for a given detection, and \textit{choose} a set of features that would allow for the maximum accuracy achieved in the cost allocated. While this strategy allows for an explicit allocation of budget towards a specific detection, the budget is not sensitive to the data itself but is chosen according to need. Our work also focuses on choosing the features based on the input scene so that a \textit{good enough} accuracy can be achieved.
Finally, it is unclear how this strategy could be extended to neural network-based methods, as it assumes a cascaded classifier structure, which is not true for many classes of networks, while our approach applies to feature-based and neural network-based classifiers, as it assumes a general structure.

\subsection{Neural Network-Based Object Detection}\label{subsec:nnbasedresults}

\begin{figure*}[t]
    \centering %
    \begin{subfigure}{0.30\textwidth}
        \includegraphics[width=\linewidth]{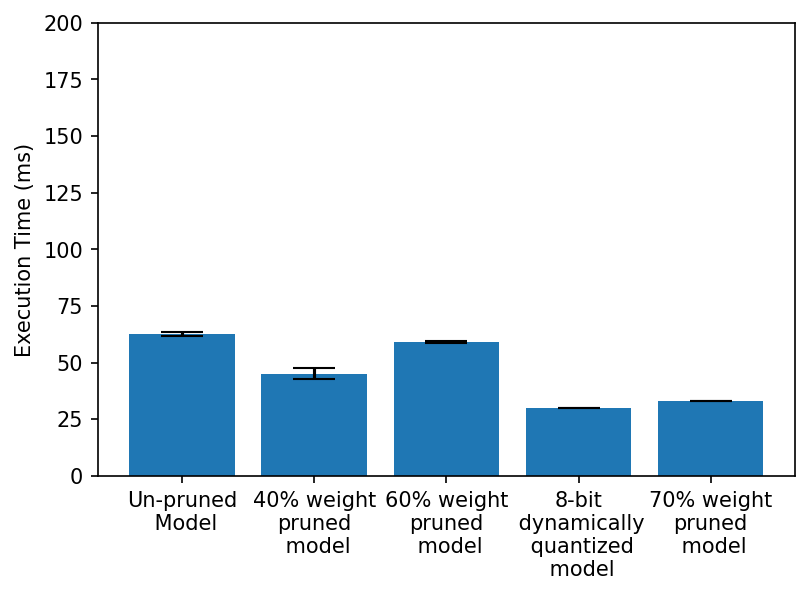}
        \caption{Execution time}\label{fig:t-urban}
    \end{subfigure}
    \begin{subfigure}{0.32\textwidth}
        \includegraphics[width=\linewidth]{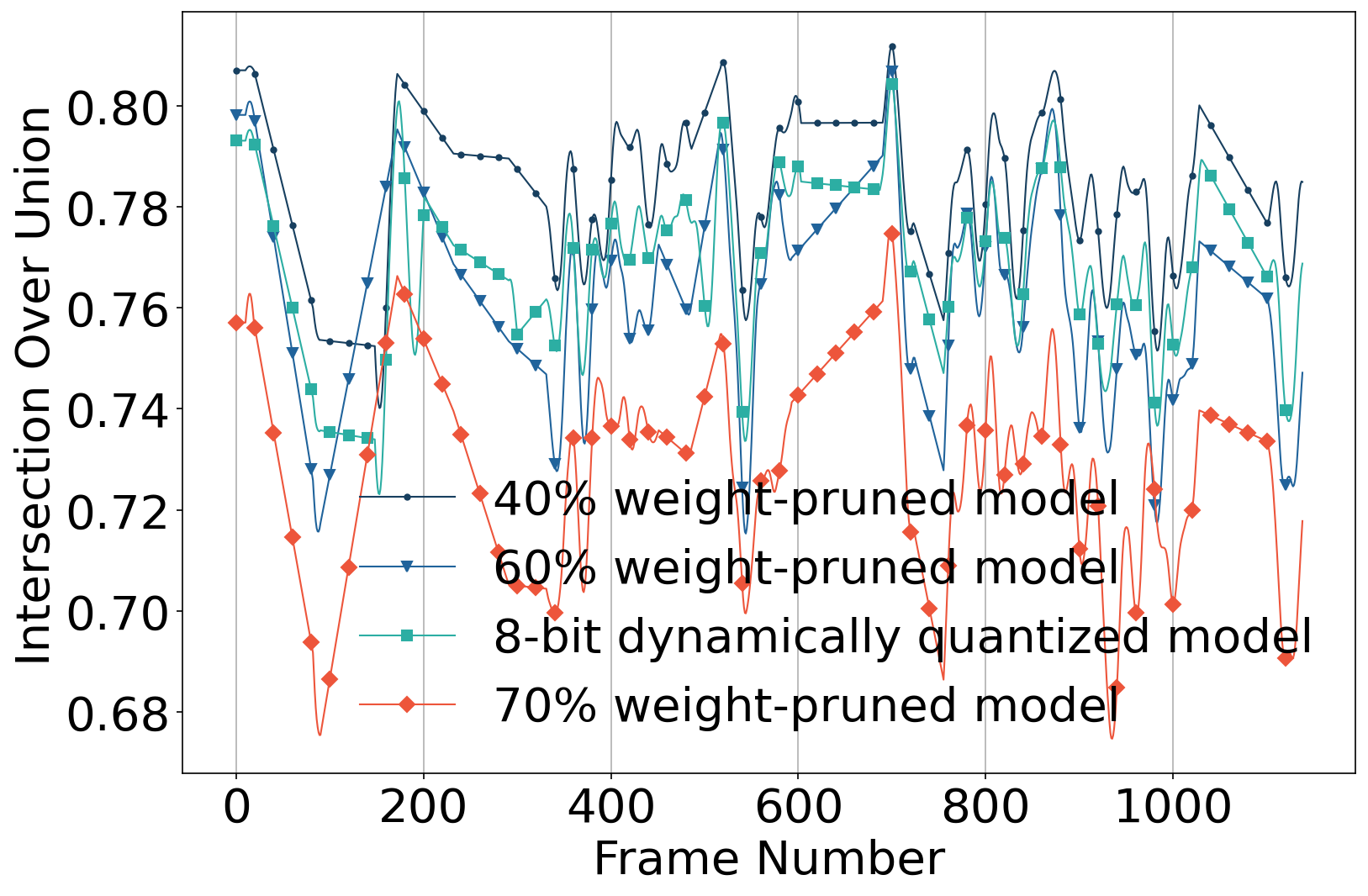}
        \caption{Intersection Over Union~(IOU)}\label{fig:iou-urban}
    \end{subfigure}
    \begin{subfigure}{0.32\textwidth}
        \includegraphics[width=\linewidth]{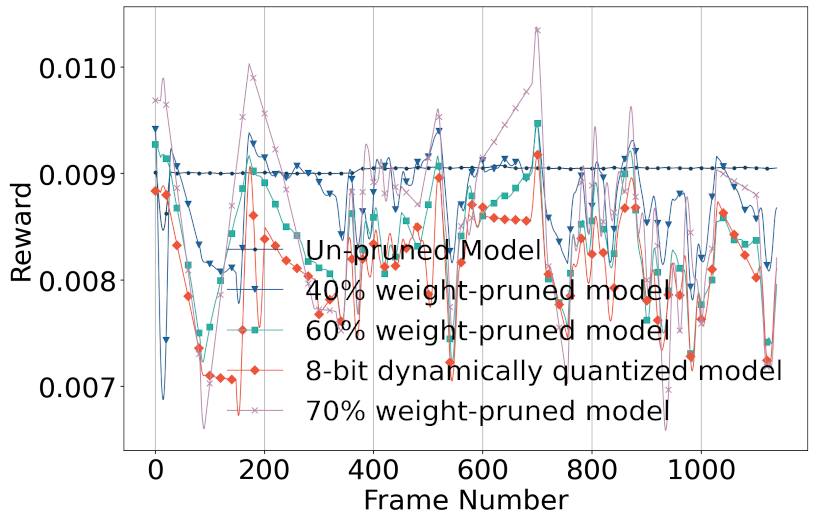}
        \caption{Reward}\label{fig:reward-urban}
    \end{subfigure}
    \begin{subfigure}{0.32\textwidth}
        \includegraphics[width=\linewidth]{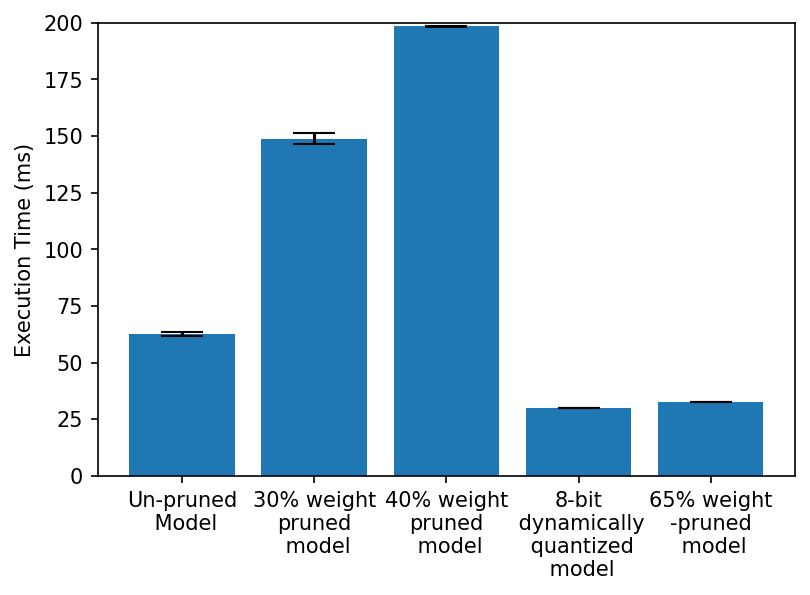}
        \caption{Execution time}\label{fig:t-suburban}
    \end{subfigure}
    \begin{subfigure}{0.32\textwidth}
        \includegraphics[width=\linewidth]{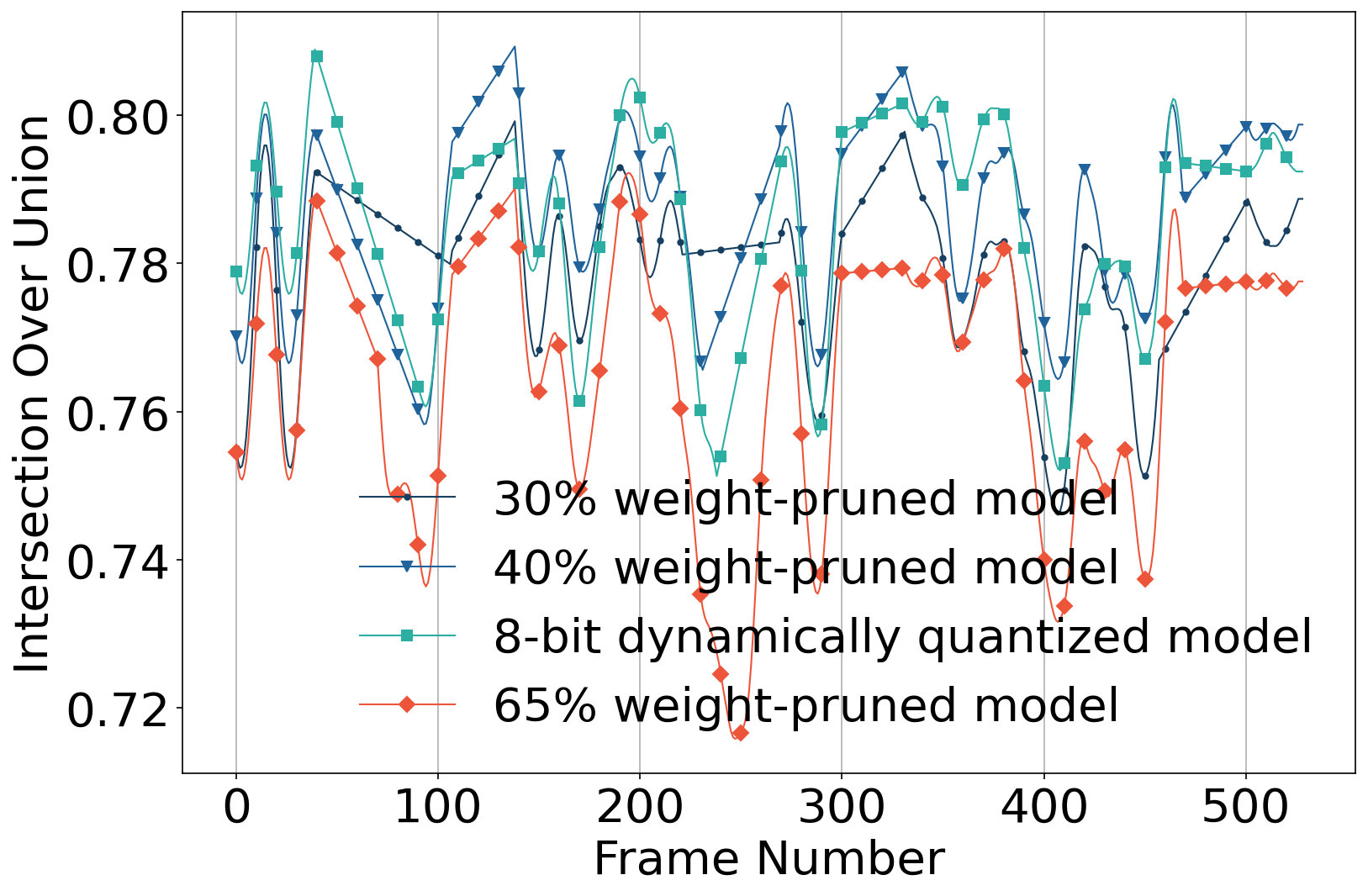}
        \caption{Intersection Over Union~(IOU)}\label{fig:iou-suburban}
    \end{subfigure}
    \begin{subfigure}{0.32\textwidth}
        \includegraphics[width=\linewidth]{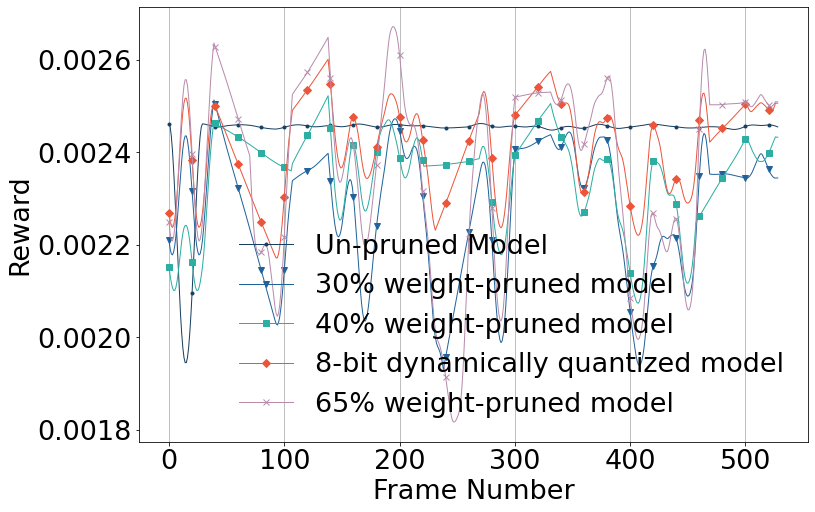}
        \caption{Reward}\label{fig:reward-suburban}
    \end{subfigure}
  \caption{Comparison among approximated models in terms of execution time, Intersection Over Union~(IOU), and reward value vs. frames---(a)-(c)~City Environment; (d)-(f)~Neighborhood Environment.}\label{fig:static-img}
\end{figure*}

\begin{table*}[t]
\small
\centering
\resizebox{\textwidth}{!}{%
\begin{tabular}{ |m{11em}|c|c|c|c|c|c| } 
\hline
\multirow{2}{*}{} & \multicolumn{2}{|c|}{Easy Maze} &
\multicolumn{2}{|c|}{Medium Maze} &
\multicolumn{2}{|c|}{Hard Maze} \\
\cline{2-7}
& Static & Our Approach & Static & Our Approach & Static & Our Approach \\ 
\hline
Average Frame-rate~[fps] & $1.9 \pm  6.4\%$ & $\mathbf{38.8 \pm 22.4\%}$ & $1.9 \pm  5.8\%$ & $\mathbf{36.5 \pm 23.6\%}$ & $2.0 \pm  4.3\%$ & $\mathbf{35.7 \pm 27.3\%}$\\ 
\hline
Time to completion~[s] & $284.9 \pm 1.4\%$ &  $\mathbf{41.2 \pm 3.0\%}$ & $604.8 \pm 1.1\%$ &  $\mathbf{100.2 \pm 2.9\%}$ & $1021.4 \pm 1.9\%$ &  $\mathbf{247.3 \pm 3.4\%}$\\ 
\hline
Average Velocity~[cm$\cdot$s$^{-1}$] & $5.3 \pm 2.3\%$ & $\mathbf{36.6 \pm 3.0\%}$ & $5.1 \pm 2.6\%$ & $\mathbf{34.6 \pm 2.7\%}$ & $4.9 \pm 2.5\%$ & $\mathbf{32.2 \pm 2.9\%}$\\ 
\hline
\end{tabular}
}
\caption{Results for the End-to-End Maze Navigation for different types of Maze.}
\label{fig:maze_results}
\end{table*}

Our tests for NN-based object detection focus on AirSim environments. First, we show the performance of our approach in two AirSim environments shipped with the software itself: City and Neighborhood environments. This performance evaluation is without any feedback control and shows the performance variation of different approximated models with respect to frames. Afterward, we expand our testing to include end-to-end navigation and real-time control in an AirSim-based custom warehouse environment, showing the viability of our approach in a real-life environment. We use Detectron2 API~\cite{wu2019detectron2} to train/test all neural networks for our experiments in this section. The details are reported in the following.

\textbf{Performance on City and Neighborhood Environments:}
As an intermediate testing stage, our framework was run on several videos from each of the City and Neighborhood environments using the TX2 GPU, and the results are shown in Fig.~\ref{fig:static-img}. The first row presents results from the City environment, while the second row presents results from the Neighborhood environment. Figs.~\ref{fig:reward-urban} and \ref{fig:reward-suburban} are of particular importance, as they show the reward estimated by the MDP for several different approximated models according to~\eqref{eq:reward}; in particular, we note that different models maximize the said rewards at different periods during a particular video clip (used for demonstration), thus justifying the use of an MDP to choose models with lower execution times depending on the input frames. This reinforces our previous similar assertion based on Fig.~\ref{fig:PerfMDP}, i.e., that a dynamic approach is better than a static approach. Similarly, Figs.~\ref{fig:iou-urban} and \ref{fig:iou-suburban} show that different models perform best for different input frames, reinforcing the point made in Fig.~\ref{fig:ApproxHog}. Similarly, the rest of the figures show the execution times of pruned and unpruned models; the higher execution times for pruned models in Fig.~\ref{fig:t-suburban} highlight the limited resources of the TX2 GPU, namely that pruned models can occasionally have higher execution times because the CPU has limited resources. This happens because the CPU has to load and unload data and neural networks from the GPU to process the incoming images. Now, given that the TX2 module has a quad-core Cortex A-57 processor with a 4-thread capacity, there is a potential for high latency with a low number of available threads.

\textbf{End-to-End Navigation in AirSim Maze Emulations:}
To validate our framework's real-world applicability, we conducted comprehensive navigation experiments across Easy, Medium, and Hard maze configurations depicted in Fig.~\ref{fig:mazes}. Our experimental setup involved deploying a drone from a predefined starting location while utilizing the emulation environment detailed in Sect.~\ref{subsec:experimentalsetup}. The system architecture enabled real-time image streaming to the TX2 GPU, which processed the data and transmitted control commands back to AirSim. To ensure realistic simulation conditions, we maintained true-to-life object dimensions (measured in centimeters) and drone response latencies (measured in seconds). %
The navigation logic leveraged the known dimensions of simulated traffic signs to compute relative distances based on detected bounding box sizes. When the drone approached within a predetermined proximity threshold of a sign, it executed appropriate turning maneuvers. We tracked completion time from initialization until the drone reached the designated endpoint. The communication overhead between AirSim and the TX2 was minimal due to the high-bandwidth ethernet connection employed. The quantitative outcomes of these trials are presented in Table~\ref{fig:maze_results}. It's worth clarifying that the reported frame rates reflect our system's processing capabilities rather than the simulator's refresh rate - demonstrating our framework's ability to handle continuous real-time visual input, even as the simulator operates at higher frequencies. The comparative analysis reveals substantial improvements over conventional static approaches, with our system achieving markedly higher frame processing rates that translate to faster goal completion times, as evidenced by the drone's enhanced average velocities. These performance gains parallel our earlier findings with HoG-based detection, demonstrating that our MDP framework effectively optimizes resource utilization across both traditional feature-based and modern neural network architectures.

\textbf{Comparison with Related Works:}
BlockDrop~\cite{wu_blockdrop_2018} uses an MDP to drop specific blocks in a Residual Network~(ResNet) to save on resources with a small loss of accuracy. For the MDP to decide on which ResNet blocks to drop, they developed a lightweight policy network for the MDP, which is trained via a reward function that penalizes loss in accuracy while rewarding using fewer blocks (less resource usage). Compared to our approach, there are some limitations that we would like to point out: i)~This paper presumes a ResNet structure for its application, while our approach is general and can be applied to both feature-based and neural network-based architectures, ii)~The impact of our approach is orthogonal to any compression techniques, as our approach just \textit{uses} these approximation techniques to create multiple alternative networks to \textit{choose} from, and finally iii)~This technique does not feature any tracking and can only be applied to still images for object detection, while our approach features an end-to-end navigation framework for robots. Furthermore, this approach only achieves a speedup of 36\% on images, as reported in the paper, while we achieve speedup as high as 42\% while maintaining 80\% accuracy.%

\section{Conclusions}\label{Sect:Conclusion}
In this work, we developed a framework that enables autonomous navigation in robots by leveraging visual cues in a manner analogous to human perception. Through extensive validation across diverse environments, we demonstrated the broad applicability of our methodology. At the core of our approach lies a Markov Decision Process (MDP) framework that dynamically tunes object detection parameters based on the input data characteristics, optimizing both computational efficiency and detection accuracy. Experimental validation on multiple datasets showed that our adaptive parameter selection reduces execution time by $20-70\%$ while maintaining high accuracy levels of $98-100\%$ compared to traditional fixed-parameter approaches.
Our hardware-in-the-loop simulations conducted across various AirSim environments, encompassing both outdoor and indoor scenarios, revealed that dynamic model selection significantly outperforms static model approaches in terms of resource utilization and accuracy metrics, yielding an $18\times$ improvement in frame processing rate. The consistent performance advantages demonstrated across different machine learning architectures underscore the versatility and effectiveness of our proposed framework. %

\textbf{Acknowledgements:}
This work was supported by the NSF RTML Award No.~1937403. A shorter version of this work appears in the Proc. of the \emph{IEEE Intl. Conference on Intelligent Robots and Systems~(IROS)}, Madrid, Spain, Oct'18~\cite{MDP_IROS}. 

\balance

\bibliographystyle{elsarticle-harv}
\bibliography{biblio/IEEEFull,biblio/IEEEConfFull,ApproxComp,CloudAssistedRobotics,biblio/cloudRobotics,biblio/objectDetection,biblio/visionMDP}

\end{document}